\documentclass{article}

\usepackage[preprint]{icml2026_fogen}

\usepackage[utf8]{inputenc}
\usepackage[T1]{fontenc}
\usepackage{hyperref}
\usepackage{url}
\usepackage{booktabs}
\usepackage{amsfonts}
\usepackage{amsmath}
\usepackage{amssymb}
\usepackage{nicefrac}
\usepackage{microtype}
\usepackage{xspace}
\usepackage{xcolor}
\usepackage[table]{xcolor}
\usepackage{graphicx}
\usepackage{subcaption}
\usepackage{multirow}
\usepackage{bbding}
\usepackage{pifont}

\definecolor{RankOne}{HTML}{DFF3E3}   
\definecolor{RankTwo}{HTML}{E6F0FA}   
\definecolor{RankThree}{HTML}{FFF1D6} 

\newcommand{\best}[1]{\cellcolor{RankOne}\textbf{#1}}
\newcommand{\second}[1]{\cellcolor{RankTwo}\underline{#1}}
\newcommand{\third}[1]{\cellcolor{RankThree}\textit{#1}}

\newcommand{\ranknote}{%
Top-1, Top-2, and Top-3 results are highlighted in green, blue, and orange, respectively.%
}

\definecolor{citecolor}{HTML}{0071bc}
\hypersetup{
    colorlinks=true,
    citecolor=citecolor,
    linkcolor=citecolor,
    urlcolor=citecolor
}

\newcommand{\ours}{Dream.exe\xspace}
\newcommand{\cmark}{\ding{51}}
\newcommand{\xmark}{\ding{55}}

\icmltitlerunning{Dream.exe: Can Video Generation Models Dream Executable Robot Manipulation?}

\begin{document}

\twocolumn[

\icmltitle{Dream.exe: Can Video Generation Models Dream Executable Robot Manipulation?}
  \icmlsetsymbol{equal}{*}
  \icmlsetsymbol{equalsecond}{\dag}

  \begin{icmlauthorlist}
    \icmlauthor{Rui Zhao}{equal,nus}
    \icmlauthor{Kaiming Yang}{equal,nus}
    \icmlauthor{Jifeng Zhu}{equalsecond,nus}
    \icmlauthor{Siyang Chen}{equalsecond,nus}
    \icmlauthor{Ziqi Wang}{nus}
    \icmlauthor{Weijia Wu}{nus} \\
    \icmlauthor{Kevin Qinghong Lin}{oxford}
    \icmlauthor{Heng Wang}{tencent}
    \icmlauthor{Mike Zheng Shou}{nus}
  \end{icmlauthorlist}

  \icmlaffiliation{nus}{Show Lab, National University of Singapore}
  \icmlaffiliation{oxford}{University of Oxford}
  \icmlaffiliation{tencent}{Tencent}

  \icmlcorrespondingauthor{Mike Zheng Shou}{mike.zheng.shou@gmail.com}

  \icmlkeywords{video generation, physical grounding, robot manipulation, benchmark, world models}

  \vskip 0.3in
]

\printAffiliationsAndNotice{\textsuperscript{*}Equal contribution. \quad \textsuperscript{\dag}Equal contribution (second authors).}

\begin{abstract}
Video generation models have made impressive strides in synthesizing visually compelling content, yet their outputs remain confined to the virtual domain.
A natural question follows: how well do these models reflect the physical world when their generated videos leave the screen and enter reality?
We propose robotic manipulation as a concrete, measurable window onto this question: if a model has truly internalized physical laws, the motion it depicts should translate into executable robot behavior.
We introduce \ours, an evaluation framework that operationalizes this criterion through a video-to-execution pipeline.
Given a scene image and a task description, \ours synthesizes a manipulation video, converts the generated motion into robot trajectories, and executes them in a physics simulator, yielding a grounding signal that purely visual metrics cannot offer.
Using this pipeline, we evaluate 8 models spanning frontier closed-source generators, open-source generators, and robot-specific models.
Our benchmark covers 101 manually curated manipulation tasks at three levels of physical complexity, measured across visual quality, trajectory fidelity, and execution success.
Encouragingly, several models achieve measurable execution success, suggesting that generative priors learned from internet-scale data already encode meaningful physical knowledge.
Yet visual quality proves a poor predictor of executability, exposing a dimension of model capability that standard visual evaluations do not capture.
\ours will be open-sourced at \url{https://github.com/showlab/Dream.exe}.
\end{abstract}

\section{Introduction}
\label{sec:intro}

Recent years have seen video generation models cross a qualitative threshold.
Models such as Wan~\cite{wan}, Kling~\cite{kling}, Imagen video~\cite{ho2022imagen}, and Veo~\cite{veo} can synthesize photorealistic videos of fluid dynamics, human motion, and complex object interactions with a fidelity that was out of reach just two years ago.
The community has begun to interpret this visual fluency as evidence of something deeper: that large-scale video generation models are learning implicit world models~\cite{sora2024,kang2024how,ha2018world}, acquiring internal representations of physical causality from the statistical regularities of internet-scale data.
This interpretation has become a foundation for an active line of research in robot learning, where generated videos are proposed as scalable behavioral priors that could reduce dependence on costly physical demonstrations~\cite{du2023unipi,jang2025dreamgen,ye2026dreamzero,liang2025videopolicy}.

The world model hypothesis, however, has never been directly tested.
Standard video generation benchmarks evaluate models on visual quality, temporal consistency, and human aesthetic ratings, all of which measure how natural a video looks without asking whether its implied motions could actually accomplish the depicted task in the physical world.
Under these metrics, a model that generates a robot arm gracefully passing through a table is indistinguishable from one whose motions are physically valid.
As models grow larger and more visually convincing, the field has no principled way to know whether their physical knowledge and learning are keeping pace.

We argue that robotic manipulation offers the right test.
The criterion is simple and unambiguous: if a model has internalized the physical laws governing a manipulation task, the trajectory implied by its generated video should produce task success when executed by a robot.
We build \ours on this intuition, treating task success in simulation as the grounding signal rather than only relying on perceptual quality scores.

\begin{figure*}[t]
  \centering
  \includegraphics[width=\linewidth]{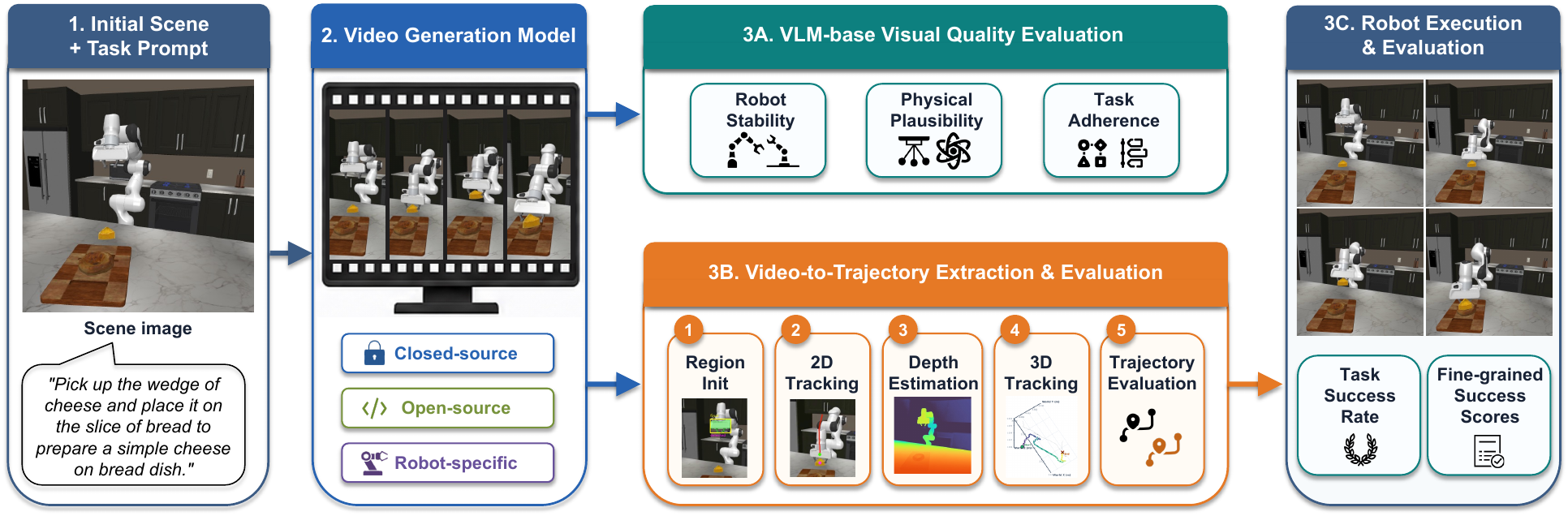}
  \caption{\textbf{The \ours\ evaluation pipeline.}
    Given an initial scene image and a task prompt,
    a video generation model produces a manipulation video.
    The video is assessed for visual quality and physical plausibility,
    and its implied motion is extracted as a robot trajectory.
    The trajectory is then executed in a physics simulator,
    where task success is the final arbiter.}
  \label{fig:pipeline}
\end{figure*}

As illustrated in Figure~\ref{fig:pipeline}, \ours\ operationalizes this criterion at scale.
Each model is given an initial scene image and a natural-language task description and asked to generate a manipulation video.
The video is then assessed along three tracks: visual evaluation of robot stability, physical plausibility, and task adherence; a five-step video-to-trajectory extraction pipeline and the corresponding trajectory evaluation; and closed-loop execution in a physics simulator that yields fine-grained success scores and an overall task success rate.

Bridging video and physical execution is non-trivial.
A generated video encodes motion only implicitly, in the form of pixel-level appearance changes, without any explicit representation of 3D geometry, contact forces, or gripper state.
To recover executable trajectories, we develop a video-to-execution pipeline that lifts 2D end-effector motion into world-frame 3D trajectories using monocular depth estimation and known camera parameters, infers gripper timing from the interaction context, and converts the result into a structured action stream that a robot controller can follow.
The task suite is built on 101 manually curated episodes from RoboCasa365~\cite{robocasa}, stratified into three difficulty levels ranging from single-object atomic manipulation to multi-stage composite tasks.
Together, these three axes of assessment provide a capability profile that no prior benchmark offers.

Using \ours, we evaluate 8 models spanning frontier closed-source generators, open-source generators, and a robot-specific policy model.
Our experiments surface three findings.
First, several models achieve measurable execution success rates, suggesting that generative priors trained on internet-scale data do encode meaningful physical knowledge.
Second, visual quality is a poor predictor of executability: models that lead on standard visual metrics frequently fail in execution, while models with modest visual scores can produce physically valid trajectories.
Third, the robot-specific policy model does not consistently outperform general generators, as the latter generalize better across diverse tasks and camera viewpoints.

\ours will be open-sourced to support future work at the intersection of video generation and robot learning. Our contributions are summarized as follows:
\begin{itemize}
  \item We introduce \ours, the first benchmark to evaluate video generation models on physical executability, using task success in simulation as the primary criterion rather than perceptual quality scores.
  \item We propose a three-track evaluation protocol: visual assessment of generated videos; video-to-trajectory extraction pipeline and trajectory evaluation; and closed-loop execution in a physics simulator and evaluation.
  \item We provide a comprehensive empirical analysis of 8 video generation models spanning closed-source, open-source, and robot-specific models, characterizing systematic failure modes and revealing that visual quality is a poor predictor of physical executability.
\end{itemize}

\section{Related Work}
\label{sec:related}

\paragraph{Video Generation Models.}
Video generation has evolved rapidly from early diffusion-based approaches into a diverse ecosystem of powerful models.
Ho et al.~\cite{ho2022vdm} established the core paradigm of applying diffusion models to video; 
Make-A-Video~\cite{singer2022makeavideo} demonstrated text-to-video generation without paired supervision;
and Stable Video Diffusion~\cite{blattmann2023svd} showed that large-scale image-to-video pretraining yields strong motion priors.
More recent open-weight models such as CogVideoX~\cite{yang2024cogvideox} and HunyuanVideo~\cite{kong2024hunyuanvideo} match or surpass earlier proprietary systems in quality and efficiency.
On the frontier, Sora~\cite{sora2024} reframed video generation as world simulation, followed by Movie Gen~\cite{polyak2024moviegen}, and the current generation of image-to-video models evaluated in this work: Kling~\cite{kling}, Wan~\cite{wan}, SeedDance~\cite{seeddance}, Veo~\cite{veo}, and LTX-Video~\cite{ltx}.
Despite their visual fluency, these models are evaluated exclusively on perceptual quality metrics; whether their generated motions are physically executable has not been tested.

\paragraph{Video Generation Benchmarks.}
Standard benchmarks evaluate video models on visual and semantic quality.
EvalCrafter~\cite{evalcrafter2024} proposes a holistic framework spanning visual quality, motion quality, and text-video alignment.
VBench~\cite{vbench2024} decomposes evaluation into fine-grained dimensions including temporal consistency, subject identity, and aesthetics.
T2V-CompBench~\cite{t2vcompbench2025} focuses on compositional reasoning over spatial relations, attributes, and actions.
These benchmarks measure how natural a video looks; they do not probe whether its physics is correct.
A growing body of work has begun to fill this gap.
VideoPhy~\cite{videophy2025} and PhyGenBench~\cite{phygenbench2025} test whether generated videos depict physically plausible phenomena, using VLM-based scorers and human raters as judges.
WorldSimBench~\cite{worldsimbench2024} adds an implicit manipulative evaluation that asks whether a video generation model could support downstream task execution via a learned policy.
MIND~\cite{ye2026mind} evaluates memory consistency and action control in world models, testing whether generated scenes remain consistent under closed-loop revisiting.
Kang et al.~\cite{kang2024how} probe model adherence to concrete physical laws and find systematic failures across all current generators.

Despite this progress, none of these works closes the loop with a real robot controller: measuring physical plausibility through visual classifiers differs categorically from asking whether a generated trajectory succeeds when executed in a physics simulator.
\ours\ makes physical executability the primary metric, directly bridging this gap.

\paragraph{Robot Learning from Video.}
The idea of using video as a source of robot behavioral knowledge spans imitation from human demonstrations and pre-training on internet-scale video~\cite{wu2024gr1}.
An influential early direction treats video generation itself as the policy: UniPi~\cite{du2023unipi} frames planning as text-conditioned video generation; SuSIE~\cite{black2024susie} synthesizes visual subgoals via image-editing diffusion models for hierarchical control; and Dreamitate~\cite{liang2024dreamitate} distills visuomotor policies directly from generated demonstrations.
The most recent wave turns video world models into zero-shot and few-shot robot policies: Cosmos Policy~\cite{cosmos} fine-tunes a video foundation model on robot demonstration data for visuomotor control; DreamGen~\cite{jang2025dreamgen} generates neural trajectories conditioned on novel environments to unlock out-of-distribution generalization; DreamZero~\cite{ye2026dreamzero} embeds action generation into the video diffusion process, achieving zero-shot policy transfer across embodiments; VideoVLA~\cite{shen2025videovla} jointly models video, language, and action to turn video generators into generalizable robot manipulators; and Video Generators are Robot Policies~\cite{liang2025videopolicy} proposes a modular framework in which a single video generator serves as the policy backbone for a wide range of manipulation skills.
Trajectory extraction methods recover executable actions from video without explicit labels: Video Prediction Policy~\cite{hu2024vpp} decodes implicit robot control signals from video diffusion representations; and Dream2Flow~\cite{dharmarajan2025dream2flow} lifts 3D object flow directly from generated videos for open-world manipulation.
Our work differs fundamentally from all of the above: \ours\ treats video generation as a \emph{fixed test subject}, evaluating the physical content of generation as-is via execution in a physics simulator built on RoboCasa365~\cite{robocasa} and robosuite~\cite{robosuite}.

\section{\ours: Benchmark Design}
\label{sec:benchmark}

\subsection{Task Suite}
\label{sec:tasks}

A benchmark for physical executability must ensure that each task scenario is strictly reproducible: the same initial scene state must be recoverable on demand, so that different video generation models can be compared on an equal footing.
We build our task suite on top of RoboCasa365~\cite{robocasa}, a large-scale simulation framework comprising 365 everyday manipulation tasks.

\begin{figure*}[h]
  \centering
  
  \includegraphics[width=\linewidth]{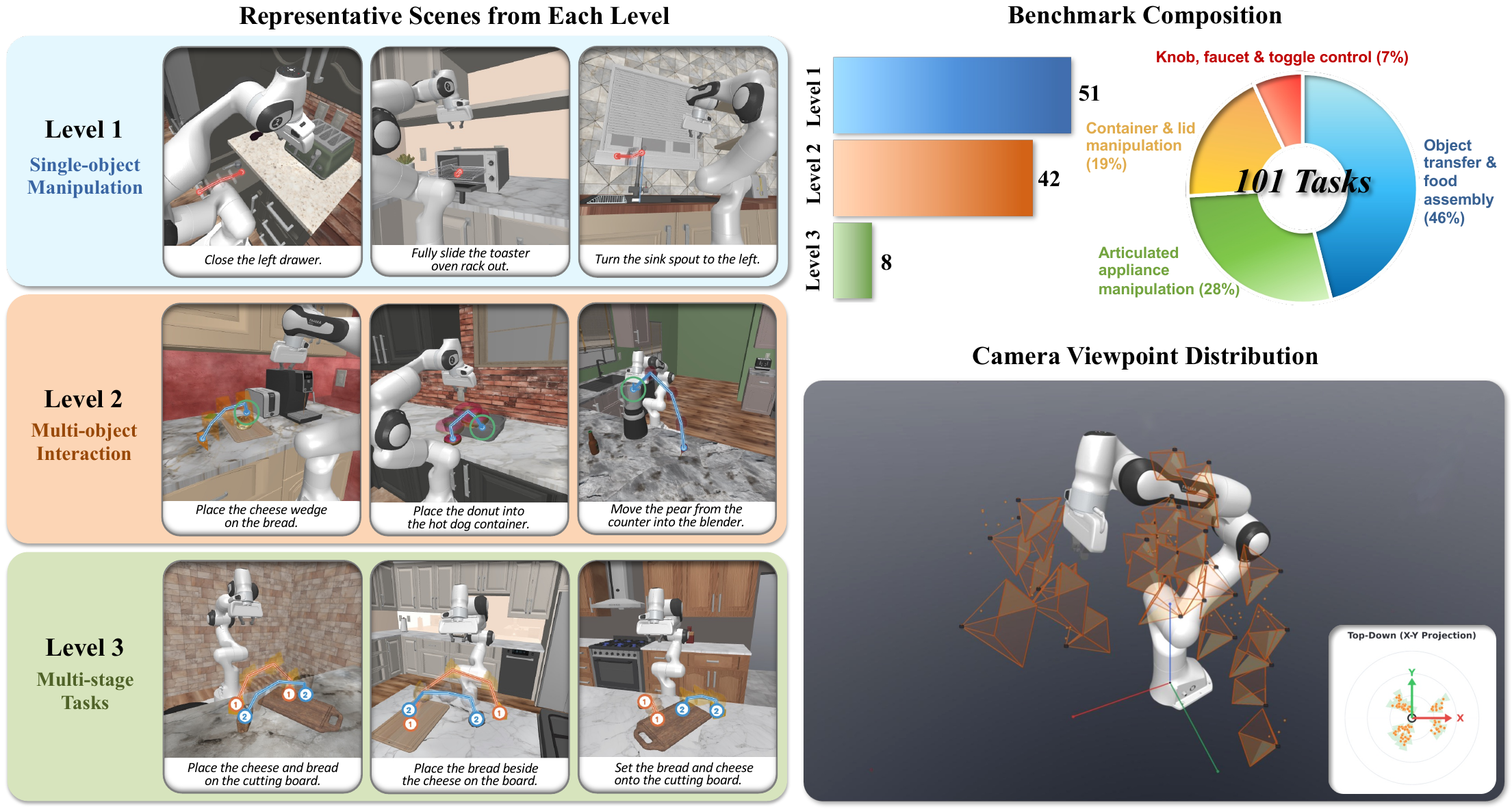}
    \caption{\textbf{Overview of the \ours\ task suite.}
    Left: representative scenes and task prompts from each difficulty level.
    Top right: distribution of 101 tasks across the three levels.
    Bottom right: camera viewpoints are deliberately diversified across scenes to improve
    generalization coverage.}
  \label{fig:tasks}
\end{figure*}

\paragraph{Data curation.}
Not all episodes are suitable for benchmarking video generation models.
Cluttered viewpoints obscure end-effector motion; ambiguous object identities make trajectory evaluation ill-defined; and certain tasks require base navigation that the current extraction pipeline does not support.
We therefore conducted a substantial manual curation effort: each candidate episode was reviewed for camera suitability, object visibility, trajectory clarity, and semantic unambiguity.
Camera viewpoints were individually tuned to maximize both object and end-effector visibility in the rendered frame.
After filtering, around 101 episodes were selected, as shown in Fig.~\ref{fig:tasks}, and are organized into a benchmark dataset with unified metadata, including the initial image and textual task prompt.

\paragraph{Three-level difficulty taxonomy.}
The tasks are stratified into three levels of increasing complexity, designed to probe different aspects of physical complexity in generated videos, as shown in Fig.~\ref{fig:tasks}.

\textit{Level 1: Atomic single-object manipulation.}
Each task involves a single object and a single continuous interaction primitive, such as pick-and-place, articulated joint actuation, button press, or knob rotation.
These tasks require the model to generate geometrically consistent end-effector motion and correct grasp-release timing, but do not demand reasoning about object-to-object relationships.

\textit{Level 2: Multi-object interaction.}
Tasks at this level involve two or more objects whose states are coupled.
Representative examples include placing one object into a container, stacking objects, or transferring contents between containers.
Success requires that the generated video correctly represent the spatial relationships between objects and the sequential dependency between manipulation events.

\textit{Level 3: Multi-stage composite tasks.}
Each task at this level decomposes into two or more semantically distinct stages, such as opening a drawer before retrieving an object, or turning a stove knob before moving a cooking vessel.
These tasks test whether a video generation model can maintain physical coherence across a long task horizon, correctly sequencing sub-goals and transitions between interactions.

\subsection{Models Evaluated}
\label{sec:models}

A central goal of \ours\ is to provide a broad and representative evaluation that spans the current landscape of video generation.
We include three categories of models, with detailed generation settings reported in Appendix~\ref{app:models}.

\paragraph{Frontier closed-source generators.}
We evaluate five state-of-the-art commercial image-to-video models: Hailuo~2.3~\cite{hailuo23} from MiniMax, Kling~3.0~\cite{kling,kling30} from Kuaishou, Wan~2.7~\cite{wan,wan27} from Alibaba, SeedDance~2.0~\cite{seeddance} from ByteDance, and Veo~3.1~\cite{veo} from Google DeepMind.
These models represent the current ceiling of general-purpose video generation quality and are the systems most commonly cited in the community.
Including them is essential for answering whether the best available generators already encode sufficient physical understanding for robot execution.

\paragraph{Open-source generators.}
We include two open-weight models: Wan~2.2~\cite{wan} and LTX-Video~\cite{ltx,ltx23}.
These models are fully reproducible and serve two purposes: they establish a baseline that the research community can build on, and they allow a controlled comparison between the open and closed variants of the same model family to isolate the effect of scale and proprietary training data. Additionally, we fine-tune Wan~2.2 on the RoboCasa365 episodes outside the test set to examine whether in-domain video data can close the domain gap between the general and robotic domains.

\paragraph{Robot-specific policy model.}
We include Cosmos~Policy~\cite{cosmos} from NVIDIA, a video generation model trained specifically on robot manipulation data.
Its inclusion directly addresses the question of whether task-specific training confers an advantage in physical executability over models trained purely on general internet video.

\subsection{Evaluation Pipeline}
\label{sec:pipeline}

\paragraph{Stage 1: Video Generation.}
Each model receives the initial scene image and the task prompt and generates a manipulation video.
For Level~1 and Level~2 tasks, a short clip is generated; for Level~3 multi-stage tasks, a longer video is generated to accommodate the extended task horizon.
Full generation settings are provided in Appendix~\ref{tab:models}.

\paragraph{Stage 2: Visual Quality Evaluation.}
Generated videos are scored before trajectory extraction to characterize visual stability, physical plausibility, and task adherence.
The additional human-evaluation protocol and results are described at the end of Section~\ref{sec:experiments}.

\paragraph{Stage 3: Video-to-Trajectory Extraction and Evaluation.}
The proposed video-to-trajectory extraction pipeline converts a manipulation video into a step-level robot action stream through a five-step chain.

\textit{Region Mask Initialization.}
On the first video frame, the module identifies the spatial region of the end-effector and the manipulated object.
When a matching simulation scene is available, initialization-time instance segmentation provides pixel masks directly.
Otherwise, open-vocabulary detection via Grounding DINO~\cite{gdino} followed by SAM2~\cite{sam2} segmentation is used to obtain the corresponding masks.

\textit{2D point tracking.}
A set of mask-based query points is sampled within each identified region and tracked across all video frames using CoTracker~\cite{cotracker}, yielding per-frame pixel coordinates and visibility flags for both the end-effector and the object.

\textit{Depth estimation and 3D lifting.}
For generated videos, video depth is estimated using the DVD~\cite{zhang2026dvd} model with LoRA adaptation on robot rollout videos. The model predicts affine depth, which is calibrated to metric scale using depth from the initial scene.
Each valid tracked pixel is transferred to a 3D point in the world frame using the camera intrinsic and extrinsic parameters associated with the scene.
Lifted trajectories are maintained in the world frame, with action deltas later emitted in the configured controller reference frame.

\textit{End-Effector Trajectory Extraction.}
A per-frame visual center is estimated from the lifted point set. 
Since the visual center of the end-effector does not directly correspond to the robotic control site, we develop a module that applies a calibration derived from the initial state to convert the visual center trajectory into the trajectory of the controller reference point.
This step is critical for physically valid execution and enables the same extraction pipeline to operate across different robot morphologies.
End-effector orientation is estimated by applying Kabsch alignment to lifted end-effector points across frames.

\textit{Gripper-Aware Action Assembly.}
The gripper open-and-close schedule is inferred from the relative motion between the end-effector trajectory and the manipulated-object trajectory.
When task annotations are available, the stage-level priors constrain the expected close/open events for each interaction type, while for multi-stage tasks, each stage is processed with its own target object and then merged into a single video-level gripper schedule.
Combining this schedule with the calibrated end-effector motion yields the executable action stream.

\paragraph{Stage 4: Robot Execution and Evaluation.}
The extracted action stream is executed in MuJoCo via the robosuite~\cite{robosuite} control framework on a Franka Panda robot.
The scene is restored to its exact initial state before each trial.
Execution proceeds in closed loop: at each checkpoint boundary, the current end-effector pose is compared to the target pose, and a correction sequence is applied if the deviation exceeds a threshold.
This prevents open-loop error accumulation and provides a controlled test of whether the trajectory extracted from the generated video can be reliably followed by the robot controller.

\section{Experiments}
\label{sec:experiments}

\paragraph{Experimental setup.}
All models are evaluated on the full task suite under a unified protocol.
For each task, the model receives the same initial scene image and natural-language prompt, while no additional context or few-shot demonstrations are provided.
We consider two instruction variants: (1) \textit{standard instructions} taken verbatim from the original dataset annotations, and (2) \textit{enhanced instructions} rephrased by a VLM, Gemini 3 Pro, into a more descriptive natural-language style that better matches the input distribution of generative models.
Each model generates a separate set of videos for each instruction variant, so every variant is evaluated end-to-end through the full video-to-execution pipeline.
Unless otherwise noted, all results reported in the main paper are the average over the two instruction variants, while the individual results under standard and enhanced instructions are provided in Appendix~\ref{app:quant}.

Several models deviate from the basic testing mode and are described below.
Wan~2.2-LoRA$_{2K}$ and Wan~2.2-LoRA$_{7K}$ are fine-tuned versions of Wan~2.2 trained on RoboCasa episodes that do not overlap with our test suite, using 2K and 7K optimization steps, respectively.
CosmosPolicy requires multi-view input by design, so we evaluate two variants: CosmosPolicy-DefaultCam follows the standard inference protocol with three camera views, while all other models receive a single task-specific view curated per scene.
To make a fairer comparison, CosmosPolicy-BenchCam replaces the primary view with a curated main camera view of \ours\  while keeping the two remaining views at their default positions.

\subsection{Visual Evaluation}
\label{sec:visual-eval}

We score each generated video with two VLM judges, Gemini 3 Pro and Qwen3-VL-Plus, along three dimensions: \textit{robot-subject stability}, \textit{physical plausibility}, and \textit{task adherence}.
For each dimension the VLM is shown sampled frames from the video together with the task prompt and produces a numeric score. 
Full prompt templates and the scoring rubric are provided in Appendix~\ref{app:vlm-eval}.
Table~\ref{tab:main} reports the results, which are the average over the two VLM judges, while the per-judge scores are provided in Appendix~\ref{app:quant}.

\begin{table}[t]
    \caption{
    \textbf{Visual quality evaluation results.}
    Results are grouped by difficulty level. Stab., Phys., and Task Adh. denote robot stability, physical plausibility, and task adherence. Higher is better ($\uparrow$).
    \ranknote{}
    }
  \label{tab:main}
  \centering
  
  \resizebox{\columnwidth}{!}{%
  \scriptsize
  \setlength{\tabcolsep}{3.5pt}
  \begin{tabular}{lcccccccccccc}
    \toprule
    \textbf{Model} & \multicolumn{3}{c}{\textbf{Level~1}} & \multicolumn{3}{c}{\textbf{Level~2}} & \multicolumn{3}{c}{\textbf{Level~3}} & \multicolumn{3}{c}{\textbf{Overall}} \\
    \cmidrule(lr){2-4}\cmidrule(lr){5-7}\cmidrule(lr){8-10}\cmidrule(lr){11-13}
    & \textbf{Stab.} & \textbf{Phys.} & \textbf{Task Adh.} & \textbf{Stab.} & \textbf{Phys.} & \textbf{Task Adh.} & \textbf{Stab.} & \textbf{Phys.} & \textbf{Task Adh.} & \textbf{Stab.} & \textbf{Phys.} & \textbf{Task Adh.} \\
    \midrule
    Hailuo~2.3 & 5.708 & 2.050 & 2.425 & 5.726 & 2.032 & 2.889 & 5.278 & 2.028 & 2.306 & 5.690 & 2.041 & 2.611 \\
    Kling~3.0 & \second{6.491} & \second{2.179} & \best{2.780} & 6.159 & 2.067 & 2.972 & 6.889 & \best{2.361} & 2.389 & 6.376 & \second{2.144} & 2.837 \\
    SeedDance~2.0 & 6.135 & \third{2.123} & \third{2.739} & 6.937 & \second{2.143} & \third{3.036} & \third{7.917} & \second{2.111} & \best{3.111} & 6.574 & \third{2.130} & \second{2.884} \\
    Veo~3.1 & 6.079 & 1.821 & \second{2.745} & 5.274 & 1.948 & \best{3.385} & 4.972 & 2.028 & \second{3.083} & 5.678 & 1.886 & \best{3.031} \\
    Wan~2.2 & 5.833 & 2.104 & 2.192 & 5.968 & 2.067 & 2.337 & 6.611 & 1.944 & 1.806 & 5.936 & 2.079 & 2.229 \\
    Wan~2.7 & \third{6.428} & 2.116 & 2.670 & \third{6.996} & \third{2.135} & \second{3.083} & 6.111 & \third{2.111} & \third{2.833} & \third{6.645} & 2.124 & \third{2.851} \\
    LTX~2.3 & 5.484 & \best{2.557} & 2.635 & 5.710 & \best{2.246} & 2.187 & 4.806 & 1.917 & 1.778 & 5.538 & \best{2.389} & 2.398 \\
    Wan 2.2-LoRA$_{2K}$ & 6.091 & 2.057 & 2.057 & 6.857 & 2.024 & 2.337 & 6.194 & 2.000 & 1.778 & 6.416 & 2.040 & 2.157 \\
    Wan 2.2-LoRA$_{7K}$ & 6.160 & 2.094 & 2.264 & 6.730 & 1.968 & 2.254 & 6.917 & 1.972 & 1.861 & 6.445 & 2.035 & 2.236 \\
    CosmosPolicy-DefaultCam & 6.417 & 2.063 & 2.143 & \second{7.518} & 2.035 & 1.930 & \second{8.111} & 1.944 & 1.750 & \second{6.881} & 2.045 & 2.047 \\
    CosmosPolicy-BenchCam & \best{7.262} & 2.020 & 2.024 & \best{7.702} & 1.921 & 1.921 & \best{8.889} & 1.944 & 1.722 & \best{7.532} & 1.985 & 1.968 \\
    \bottomrule
  \end{tabular}}
\end{table}

CosmosPolicy-BenchCam scores highest on robot-subject stability, consistent with its domain-specific training on robotic footage.
Veo~3.1 leads on task adherence and LTX~2.3 on physical plausibility.
To complement these automatic scores and mitigate the uncertainty inherent in black-box VLM evaluation, we also conduct a human study with the same dimensions, reported in Section~\ref{sec:human-eval}.

\subsection{Video-to-Trajectory Evaluation}
\label{sec:v2t-eval}

As reported in Table~\ref{tab:traj-metrics-union-all-models}, we compare extracted 3D trajectories against ground-truth rollout trajectories with three metrics.
HSD is the symmetric Hausdorff distance computed on the most spatially extended sub-trajectory of the ground truth, capturing worst-case shape deviation.
DYN measures the Wasserstein-1 distance between the per-frame speed distributions of the generated and reference trajectories, reflecting how closely the motion dynamics are reproduced.
NDTW is the DTW alignment cost divided by the alignment path length, penalising local temporal mismatches.
All three raw distances are divided by a per-task normalization factor derived from the spatial extent and speed scale of the ground-truth trajectory, then mapped to a $[0,1]$ similarity score where higher is better.
Metrics are computed separately for the end-effector visual center, the end-effector tool center point, and the manipulated object.

\begin{table}[t]
  \caption{
    \textbf{Trajectory evaluation results.}
    EEF vis, EEF tcp, and OBJ are the end-effector visual center, end-effector tool center point, and manipulated object. HSD, DYN, and NDTW measure trajectory shape, dynamics, and temporal-alignment similarity. Higher is better ($\uparrow$).
    }
  \label{tab:traj-metrics-union-all-models}
  \centering
  \resizebox{\columnwidth}{!}{%
  \begin{tabular}{lccccccccc}
    \toprule
    \textbf{Model}
    & \multicolumn{3}{c}{\textbf{EEF vis}}
    & \multicolumn{3}{c}{\textbf{EEF tcp}}
    & \multicolumn{3}{c}{\textbf{OBJ}} \\
    \cmidrule(lr){2-4}\cmidrule(lr){5-7}\cmidrule(lr){8-10}
    & \textbf{HSD} & \textbf{DYN} & \textbf{NDTW}
    & \textbf{HSD} & \textbf{DYN} & \textbf{NDTW}
    & \textbf{HSD} & \textbf{DYN} & \textbf{NDTW} \\
    \midrule
    Hailuo~2.3 & 0.623 & 0.638 & 0.704 & 0.716 & 0.715 & 0.822 & 0.555 & 0.821 & 0.720 \\
    Kling~3.0 & \third{0.733} & 0.740 & \second{0.831} & \third{0.734} & 0.740 & \second{0.836} & 0.517 & 0.766 & 0.689 \\
    SeedDance~2.0 & 0.692 & 0.698 & 0.796 & 0.700 & 0.704 & 0.816 & 0.485 & 0.751 & 0.659 \\
    Veo~3.1 & 0.537 & 0.564 & 0.601 & 0.716 & 0.754 & 0.812 & 0.526 & 0.789 & 0.675 \\
    Wan~2.2 & 0.639 & 0.708 & 0.727 & 0.663 & 0.738 & 0.764 & 0.505 & 0.717 & 0.588 \\
    Wan~2.7 & \second{0.753} & \second{0.778} & \best{0.838} & \second{0.762} & \second{0.789} & \best{0.862} & \third{0.599} & \second{0.852} & \second{0.750} \\
    LTX~2.3 & 0.569 & 0.627 & 0.623 & 0.710 & \third{0.782} & 0.797 & \second{0.602} & \third{0.838} & \third{0.742} \\
    Wan 2.2-LoRA$_{2K}$ & 0.645 & 0.717 & 0.736 & 0.669 & 0.740 & 0.769 & 0.498 & 0.696 & 0.581 \\
    Wan 2.2-LoRA$_{7K}$ & 0.671 & \third{0.754} & 0.766 & 0.677 & 0.765 & 0.780 & 0.466 & 0.700 & 0.526 \\
    CosmosPolicy-BenchCam & \best{0.770} & \best{0.833} & \third{0.823} & \best{0.770} & \best{0.839} & \third{0.835} & \best{0.629} & \best{0.873} & \best{0.798} \\
    \bottomrule
  \end{tabular}
  }
\end{table}

Wan~2.7 leads on or is competitive on end-effector trajectory similarity, while CosmosPolicy-BenchCam leads on object trajectory similarity.
Notably, general-purpose models such as Wan~2.7 and Kling~3.0 match or exceed CosmosPolicy on several end-effector metrics, suggesting that large-scale pretraining on general video can rival robot-specific training in terms of generating suitable robot trajectories.

\subsection{Robot Execution Evaluation}
\label{sec:exec-eval}

The extracted trajectories are executed in the corresponding robosuite simulation environments and evaluated at two levels.
Table~\ref{tab:exec_metrics_by_level} reports trajectory executability metrics that measure how easily the video-implied trajectory can be realized by the robot controller: E-SR is the fraction of intermediate checkpoints reached, nDTW measures dense TCP tracking disagreement between the commanded and executed trajectories, Pos95 and Rot95 are 95th-percentile position and rotation errors, and Smth is path-normalized executed smoothness.
Table~\ref{bench-exec-outcomes} reports task-level execution evaluation results, measuring whether the robot actually completes the manipulation task. SR-B is the binary success rate and SR-P is a continuous 0--1 progress score that remains informative even when SR-B is zero. The sub-goal columns Rel, Place, Art, and Core measure end-effector release quality, target placement proximity, articulation completion degree, and core sub-goal fraction respectively, while their availability depends on task category and difficulty level.

\begin{table*}[t]
  \centering
  \caption{\textbf{Trajectory executability evaluation results.} Results are broken down by difficulty level and overall. E-SR is strict checkpoint executability, where higher is better ($\uparrow$). nDTW is commanded-vs-executed TCP tracking disagreement, Pos95 and Rot95 are 95th-percentile position and rotation tracking errors in cm and degrees, and Smth is $10^3{\times}$ path-normalized executed smoothness, where lower is better ($\downarrow$). }
  \label{tab:exec_metrics_by_level}
  \scriptsize
  \setlength{\tabcolsep}{1.0pt}
  \renewcommand{\arraystretch}{0.94}
  \resizebox{\textwidth}{!}{%
  \begin{tabular}{@{}l*{20}{c}@{}}
    \toprule
    \textbf{Model}
      & \multicolumn{5}{c}{\textbf{Level 1}}
      & \multicolumn{5}{c}{\textbf{Level 2}}
      & \multicolumn{5}{c}{\textbf{Level 3}}
      & \multicolumn{5}{c}{\textbf{Overall}} \\
    \cmidrule(lr){2-6}\cmidrule(lr){7-11}\cmidrule(lr){12-16}\cmidrule(lr){17-21}
      & \textbf{E-SR}$\uparrow$ & \textbf{nDTW}$\downarrow$ & \textbf{Pos95}$\downarrow$ & \textbf{Rot95}$\downarrow$ & \textbf{Smth}$\downarrow$
      & \textbf{E-SR}$\uparrow$ & \textbf{nDTW}$\downarrow$ & \textbf{Pos95}$\downarrow$ & \textbf{Rot95}$\downarrow$ & \textbf{Smth}$\downarrow$
      & \textbf{E-SR}$\uparrow$ & \textbf{nDTW}$\downarrow$ & \textbf{Pos95}$\downarrow$ & \textbf{Rot95}$\downarrow$ & \textbf{Smth}$\downarrow$
      & \textbf{E-SR}$\uparrow$ & \textbf{nDTW}$\downarrow$ & \textbf{Pos95}$\downarrow$ & \textbf{Rot95}$\downarrow$ & \textbf{Smth}$\downarrow$ \\
    \midrule
    Hailuo 2.3 & 0.508 & 26.247 & 53.096 & 25.658 & 16.865 & 0.510 & 251.421 & 833.866 & 22.175 & 17.753 & \second{0.689} & 6.552 & 2.364 & 13.272 & 18.354 & 0.519 & 118.714 & 374.759 & 23.469 & 17.323 \\
    Kling 3.0 & 0.421 & 8.964 & 9.304 & 28.324 & 18.216 & 0.514 & 23.765 & 300.346 & 27.538 & 17.445 & 0.607 & \third{3.665} & \third{2.180} & 10.638 & 19.883 & 0.470 & 14.804 & 129.908 & 26.900 & 17.995 \\
    SeedDance 2.0 & 0.437 & 20.302 & 135.486 & 28.916 & 19.579 & 0.558 & 11.193 & 65.576 & 15.356 & 20.273 & 0.604 & 5.241 & 5.431 & 12.006 & \third{18.121} & 0.497 & 15.619 & 98.689 & 22.351 & 19.781 \\
    Veo 3.1 & \third{0.522} & 28.826 & 96.989 & 24.300 & 16.142 & 0.513 & 9.098 & 8.728 & \third{12.301} & 19.906 & 0.631 & 7.466 & 4.138 & 10.658 & \second{17.648} & 0.527 & 19.248 & 54.632 & \third{17.823} & 17.821 \\
    Wan 2.2 & 0.448 & 112.789 & 918.201 & 21.013 & \third{14.066} & 0.472 & \third{8.235} & \third{7.661} & 23.045 & \second{15.594} & 0.534 & 4.078 & \best{2.119} & 12.099 & 51.818 & 0.463 & 62.853 & 485.140 & 21.282 & 16.944 \\
    Wan 2.7 & 0.513 & 8.965 & 9.056 & 24.398 & 19.206 & \second{0.617} & 39.153 & 141.321 & 18.069 & 17.519 & 0.616 & 5.518 & 6.553 & 15.151 & 18.718 & \third{0.562} & 21.314 & 63.909 & 21.249 & 18.476 \\
    LTX 2.3 & 0.422 & 9.789 & 9.817 & 20.798 & 15.255 & 0.392 & 11.315 & 9.501 & 21.148 & 20.762 & 0.252 & 16.515 & 3.205 & 23.584 & 62.001 & 0.401 & 10.813 & 9.391 & 21.225 & 19.931 \\
    Wan 2.2-LoRA$_{2K}$ & 0.465 & \third{7.707} & \third{7.704} & \third{17.991} & 14.247 & 0.464 & 8.810 & 17.107 & 19.332 & \best{15.069} & 0.612 & \second{3.158} & \second{2.156} & \third{7.927} & 52.913 & 0.474 & \third{7.895} & 11.285 & 17.907 & \third{16.886} \\
    Wan 2.2-LoRA$_{7K}$ & 0.471 & 8.889 & 8.497 & 18.952 & \second{13.668} & 0.445 & 8.958 & 8.607 & 27.265 & \third{15.717} & 0.553 & 3.815 & 2.561 & 9.811 & 53.702 & 0.465 & 8.613 & \third{8.187} & 21.788 & 16.930 \\
    CosmosPolicy-DefaultCam & \best{0.662} & \best{4.376} & \best{3.928} & \best{4.949} & \best{12.538} & \best{0.841} & \best{2.905} & \best{3.372} & \best{4.127} & 17.354 & \best{0.891} & \best{2.794} & 3.170 & \best{2.319} & 18.464 & \best{0.750} & \best{3.670} & \best{3.652} & \best{4.451} & \best{14.893} \\
    CosmosPolicy-BenchCam & \second{0.627} & \second{4.639} & \second{4.355} & \second{6.261} & 14.306 & \third{0.563} & \second{4.827} & \second{5.124} & \second{5.474} & 18.025 & \third{0.662} & 4.098 & 4.685 & \second{4.038} & \best{17.518} & \second{0.603} & \second{4.685} & \second{4.695} & \second{5.802} & \second{16.044} \\
    \bottomrule
  \end{tabular}
  }

\end{table*}

\begin{table*}[t]
  \centering
  \caption{\textbf{Task-level execution evaluation results.} SR-B is the binary task success rate and SR-P is a continuous partial-completion score. Rel, Place, Art, and Core report sub-goal completion for end-effector release, target placement, articulation progress, and core sub-goal fraction, whose availability depends on the task category and difficulty. Higher is better ($\uparrow$). }
  \label{bench-exec-outcomes}
  \scriptsize
  \setlength{\tabcolsep}{2.0pt}
  \renewcommand{\arraystretch}{0.94}
  \resizebox{\textwidth}{!}{%
  \begin{tabular}{@{}l*{20}{c}@{}}
    \toprule
    \textbf{Model}
      & \multicolumn{3}{c}{\textbf{Level 1}}
      & \multicolumn{6}{c}{\textbf{Level 2}}
      & \multicolumn{5}{c}{\textbf{Level 3}}
      & \multicolumn{6}{c}{\textbf{Overall}} \\
    \cmidrule(lr){2-4} \cmidrule(lr){5-10} \cmidrule(lr){11-15} \cmidrule(lr){16-21}
      & \textbf{SR-B}$\uparrow$ & \textbf{SR-P}$\uparrow$ & \textbf{Art}$\uparrow$
      & \textbf{SR-B}$\uparrow$ & \textbf{SR-P}$\uparrow$ & \textbf{Rel}$\uparrow$ & \textbf{Place}$\uparrow$ & \textbf{Art}$\uparrow$ & \textbf{Core}$\uparrow$
      & \textbf{SR-B}$\uparrow$ & \textbf{SR-P}$\uparrow$ & \textbf{Rel}$\uparrow$ & \textbf{Place}$\uparrow$ & \textbf{Core}$\uparrow$
      & \textbf{SR-B}$\uparrow$ & \textbf{SR-P}$\uparrow$ & \textbf{Rel}$\uparrow$ & \textbf{Place}$\uparrow$ & \textbf{Art}$\uparrow$ & \textbf{Core}$\uparrow$ \\
    \midrule
    Hailuo 2.3 & 0.104 & 0.230 & \third{0.197} & \third{0.143} & 0.592 & 0.778 & 0.305 & 0.751 & \second{0.188} & 0.000 & \second{0.359} & \best{0.688} & \third{0.031} & \third{0.031} & 0.112 & 0.387 & 0.763 & \third{0.251} & \third{0.304} & \third{0.156} \\
    Kling 3.0 & 0.123 & \third{0.270} & \best{0.230} & \second{0.190} & 0.607 & 0.547 & \best{0.463} & 0.754 & \best{0.352} & \best{0.062} & 0.297 & 0.438 & \best{0.156} & \best{0.156} & \second{0.146} & \third{0.409} & 0.529 & \best{0.402} & \best{0.331} & \best{0.312} \\
    SeedDance 2.0 & \third{0.151} & \best{0.283} & \second{0.216} & \best{0.214} & \second{0.656} & 0.815 & 0.298 & 0.759 & \second{0.188} & 0.000 & 0.328 & \second{0.625} & \third{0.031} & \third{0.031} & \best{0.165} & \best{0.439} & 0.785 & 0.244 & \second{0.320} & \third{0.156} \\
    Veo 3.1 & 0.033 & 0.105 & 0.087 & 0.120 & \third{0.611} & \second{0.882} & 0.278 & \third{0.764} & 0.143 & 0.000 & 0.266 & \third{0.500} & \third{0.031} & \third{0.031} & 0.069 & 0.345 & \third{0.820} & 0.228 & 0.228 & 0.120 \\
    Wan 2.2 & 0.038 & 0.132 & 0.076 & 0.060 & 0.509 & 0.587 & 0.290 & \second{0.773} & 0.156 & 0.000 & 0.188 & 0.375 & 0.000 & 0.000 & 0.044 & 0.290 & 0.553 & 0.232 & 0.210 & 0.125 \\
    Wan 2.7 & 0.094 & 0.215 & 0.168 & \best{0.214} & \best{0.667} & \best{0.884} & \second{0.325} & 0.760 & \second{0.188} & 0.000 & \best{0.375} & \best{0.688} & \second{0.062} & \second{0.062} & \third{0.136} & \second{0.412} & \best{0.853} & \second{0.272} & 0.282 & \second{0.163} \\
    LTX 2.3 & 0.047 & 0.140 & 0.100 & 0.037 & 0.503 & 0.712 & 0.293 & 0.722 & 0.154 & 0.000 & 0.250 & \third{0.500} & 0.000 & 0.000 & 0.039 & 0.294 & 0.678 & 0.233 & 0.220 & 0.122 \\
    Wan 2.2-LoRA$_{2K}$ & 0.038 & 0.122 & 0.079 & 0.071 & 0.500 & 0.545 & 0.302 & 0.763 & \third{0.172} & 0.000 & 0.219 & 0.438 & 0.000 & 0.000 & 0.049 & 0.284 & 0.528 & 0.241 & 0.210 & 0.138 \\
    Wan 2.2-LoRA$_{7K}$ & 0.029 & 0.144 & 0.090 & 0.071 & 0.517 & 0.595 & 0.286 & \best{0.799} & 0.156 & 0.000 & 0.219 & 0.438 & 0.000 & 0.000 & 0.044 & 0.303 & 0.570 & 0.229 & 0.226 & 0.125 \\
    CosmosPolicy-DefaultCam & \second{0.179} & 0.241 & 0.186 & 0.024 & 0.534 & 0.597 & \third{0.307} & 0.749 & \third{0.172} & 0.000 & 0.250 & \third{0.500} & 0.000 & 0.000 & 0.102 & 0.361 & 0.581 & 0.246 & 0.294 & 0.138 \\
    CosmosPolicy-BenchCam & \best{0.208} & \second{0.271} & 0.188 & 0.000 & 0.594 & \third{0.849} & 0.292 & 0.708 & 0.156 & 0.000 & \third{0.344} & \best{0.688} & 0.000 & 0.000 & 0.107 & 0.408 & \second{0.823} & 0.234 & 0.288 & 0.125 \\
    \midrule
    \rowcolor{gray!10}
    Rollout Video$^\dagger$ & 0.765 & 0.851 & 0.818 & 0.381 & 0.742 & 0.811 & 0.562 & 0.755 & 0.516 & 0.750 & 0.938 & 1.000 & 0.875 & 0.875 & 0.604 & 0.812 & 0.842 & 0.625 & 0.805 & 0.588 \\
    \rowcolor{gray!18}
    Rollout Video w/ GT Depth$^\ddagger$ & 1.000 & 1.000 & 0.950 & 0.952 & 0.979 & 0.905 & 0.866 & 1.000 & 0.953 & 1.000 & 1.000 & 1.000 & 1.000 & 1.000 & 0.981 & 0.991 & 0.920 & 0.893 & 0.960 & 0.963 \\
    \bottomrule
    \end{tabular}
  }
  \vspace{2pt}
  \begin{minipage}{0.97\textwidth}
    \scriptsize
    \emph{Note.} $^\dagger$ Rollout Video uses the same depth estimation pipeline as generated videos, while $^\ddagger$ Rollout Video (w/ GT Depth) replaces estimated depth with simulator depth. These rows serve as reference bounds for the video-to-execution pipeline. Rankings are computed among generation models only; oracle/reference rows are shaded in gray. 
  \end{minipage}
  
  \vspace{-1em}
\end{table*}

Trajectory executability metrics show consistent trends across models, where overall E-SR ranges from 0.40 to 0.75, with the robot-specific CosmosPolicy variants reaching the highest values, while nDTW, positional, and rotational errors quantify how faithfully the extracted trajectories can be followed by the robot controller.
Task-level execution results reveal a clear difficulty gradient. At Level~1, CosmosPolicy-BenchCam achieves the highest SR-B of 20.8\%, and articulation sub-goal scores vary noticeably across models.
At Level~2, SeedDance~2.0 and Wan~2.7 lead with 21.4\% SR-B. Rel scores are generally high across several models, indicating that end-effector release is reliably achieved, while Place and Core scores are more discriminative.
At Level~3, only Kling~3.0 achieves non-zero task success with 6.2\% SR-B, while most generation models remain at zero. Nevertheless, non-zero sub-goal scores indicate partial progress on multi-step tasks after execution.

It is worth emphasizing that CosmosPolicy outputs robot actions, whereas the general-purpose video generators obtain their actions through our proposed video-to-trajectory pipeline. Even under this indirect route, the general generators reach task-level SR-B that is comparable to or even exceeds CosmosPolicy, reflecting their stronger generalization across tasks and camera viewpoints.
The rollout-video reference rows further contextualize these results: replacing estimated depth with simulator ground-truth depth produces a further improvement, indicating that depth estimation remains a bottleneck in the pipeline. Crucially, this bottleneck applies uniformly to all general-purpose generators, ensuring a fair comparison across them.

\subsection{Human Evaluation}
\label{sec:human-eval}

Four independent human annotators rated each generated video on a 1--5 scale across four dimensions: robot stability, physical plausibility, task adherence, and expected execution result.
The rating results are shown in Table~\ref{tab:human-evaluation-results}.

\begin{table*}[t]
  \caption{
    \textbf{Human evaluation results.}
    Stab., Phys., Task Adh., and Exec are annotator ratings of robot stability, physical plausibility, task adherence, and expected execution result on a 1--5 scale. Higher is better ($\uparrow$).
    \ranknote{}
  }
  \label{tab:human-evaluation-results}
  \centering
  \scriptsize
  \setlength{\tabcolsep}{2.0pt} 
  \begin{tabular*}{\linewidth}{@{\extracolsep{\fill}} l *{16}{c}@{}}
    \toprule
    \textbf{Model} & \multicolumn{4}{c}{\textbf{Level~1}} & \multicolumn{4}{c}{\textbf{Level~2}} & \multicolumn{4}{c}{\textbf{Level~3}} & \multicolumn{4}{c}{\textbf{Overall}} \\
    \cmidrule(lr){2-5}\cmidrule(lr){6-9}\cmidrule(lr){10-13}\cmidrule(lr){14-17}
    & \textbf{Stab.} & \textbf{Phys.} & \textbf{Task Adh.} & \textbf{Exec} & \textbf{Stab.} & \textbf{Phys.} & \textbf{Task Adh.} & \textbf{Exec} & \textbf{Stab.} & \textbf{Phys.} & \textbf{Task Adh.} & \textbf{Exec} & \textbf{Stab.} & \textbf{Phys.} & \textbf{Task Adh.} & \textbf{Exec} \\
    \midrule
    Hailuo~2.3 & 3.192 & 2.346 & 3.500 & \third{2.423} & 3.045 & 2.500 & 3.864 & 1.864 & 2.000 & 1.500 & 3.000 & 1.250 & 2.900 & 2.233 & 3.533 & 1.983 \\
    Kling~3.0 & \third{3.885} & \third{3.269} & \best{4.769} & \second{2.462} & 3.773 & 3.909 & \second{4.591} & \best{2.955} & \second{4.083} & 2.833 & \third{4.167} & \best{3.083} & \third{3.883} & 3.417 & \best{4.583} & \best{2.767} \\
    SeedDance~2.0 & 3.538 & \second{3.462} & \second{4.423} & 2.346 & \second{4.000} & \second{4.045} & \best{4.682} & \third{2.273} & \third{3.833} & \second{3.917} & \second{4.583} & \second{2.750} & 3.767 & \second{3.767} & \second{4.550} & \third{2.400} \\
    Veo~3.1 & 2.692 & 2.423 & 3.346 & 1.654 & 2.773 & 3.182 & 4.227 & 1.591 & 2.400 & 2.300 & 4.000 & 2.300 & 2.672 & 2.690 & 3.793 & 1.741 \\
    Wan~2.2 & 3.115 & 2.615 & 2.308 & 1.577 & 3.273 & 2.773 & 2.682 & 1.773 & 2.500 & 1.900 & 1.800 & 1.300 & 3.069 & 2.552 & 2.362 & 1.603 \\
    Wan~2.7 & \second{4.000} & 3.192 & \third{4.154} & 2.269 & 4.000 & \third{3.955} & \third{4.545} & \second{2.773} & 3.500 & \third{3.900} & \best{4.800} & \third{2.600} & \second{3.914} & \third{3.603} & \third{4.414} & \second{2.517} \\
    LTX~2.3 & 2.042 & 1.625 & 2.250 & 1.667 & 1.722 & 1.500 & 1.889 & 1.111 & 1.300 & 1.000 & 1.300 & 1.000 & 1.788 & 1.462 & 1.942 & 1.346 \\
    Wan 2.2-LoRA$_{2K}$ & 3.318 & 2.682 & 2.545 & 1.636 & 3.889 & 3.333 & 2.778 & 1.889 & 3.000 & 2.600 & 1.500 & 1.300 & 3.460 & 2.900 & 2.420 & 1.660 \\
    Wan 2.2-LoRA$_{7K}$ & 3.400 & 2.850 & 2.750 & 1.800 & \third{4.000} & 3.278 & 2.833 & 2.056 & 3.100 & 2.300 & 2.000 & 1.200 & 3.562 & 2.896 & 2.625 & 1.771 \\
    CosmosPolicy-DefaultCam & \best{4.343} & \best{4.100} & 2.286 & \best{2.476} & \best{4.426} & \best{4.370} & 1.870 & 1.906 & \best{4.900} & \best{4.700} & 1.700 & 1.500 & \best{4.418} & \best{4.254} & 2.075 & 2.171 \\
    CosmosPolicy-BenchCam & 3.343 & 3.108 & 2.186 & 2.098 & 3.688 & 3.422 & 1.531 & 1.344 & 3.333 & 2.333 & 1.083 & 1.083 & 3.466 & 3.169 & 1.876 & 1.715 \\
    \bottomrule
  \end{tabular*}
\end{table*}

\begin{figure*}[h]
  \centering
  \includegraphics[width=\linewidth]{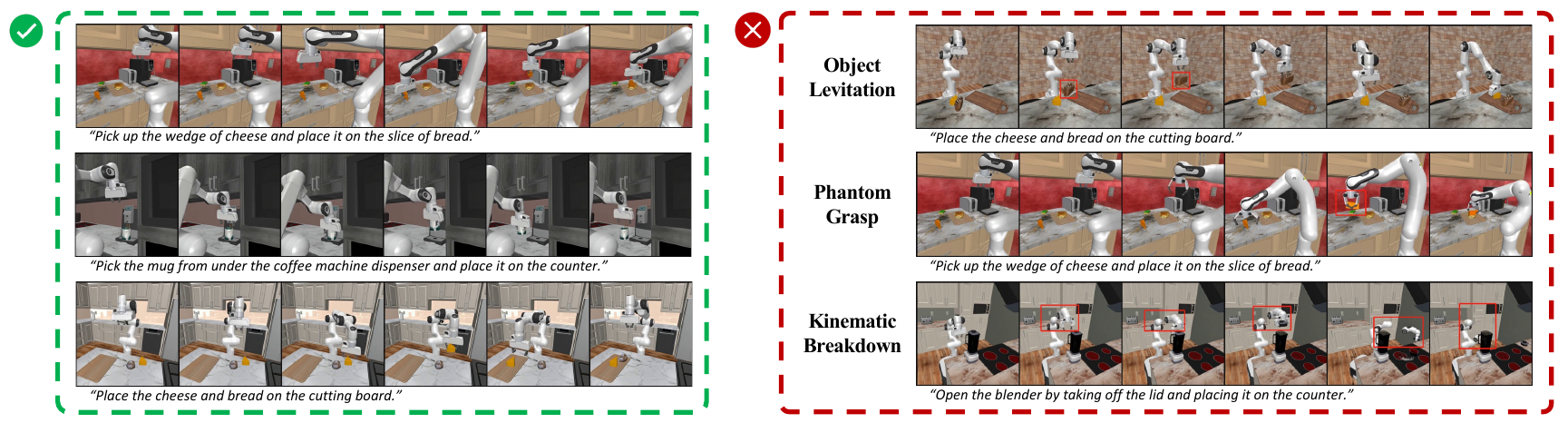}
  \caption{\textbf{Success and failure mode taxonomy.} We provide representative examples for each failure category.}
  \vspace{-1em}
  \label{fig:failures}
\end{figure*}

Among general-purpose video generators, Wan~2.7 receives the highest stability rating and SeedDance~2.0 the highest physical-plausibility rating, while Kling~3.0 leads on task adherence and expected execution result.
CosmosPolicy variants score high on stability and physical plausibility but low on task adherence and expected execution result, consistent with their tendency to produce visually robotic motion without completing the specified task.

\subsection{More Findings}
\label{sec:findings}

\paragraph{Visual quality does not equal executability.}
Visual quality is an unreliable predictor of executability.
Physical plausibility, the dimension most tied to physical correctness, is essentially uncorrelated with task success across Tables~\ref{tab:main} and~\ref{bench-exec-outcomes}, with a Pearson correlation of $r=-0.03$ against SR-B.
The mismatch is stark per model: LTX~2.3 ranks first on physical plausibility yet last on SR-B, while Veo~3.1 leads on task adherence yet reaches only $3.3\%$ Level-1 success.
Conversely, visually weaker models such as SeedDance~2.0 and Kling~3.0 achieve the strongest task-level outcomes.
Human evaluation confirms the same pattern.

\paragraph{Generative priors help, but struggle at long horizons.}

Several general-purpose models achieve non-trivial task success without any robot-specific supervision: SeedDance~2.0 reaches 15.1\% SR at Level~1, while SeedDance~2.0 and Wan~2.7 reach 21.4\% SR at Level~2.
However, Level~3 remains difficult, where only Kling~3.0 achieves non-zero task success, and most models fail to complete.

\paragraph{Robot-specific training sharpens geometry more than task success.}
CosmosPolicy leads on checkpoint executability at Levels~1--2, yet falls substantially behind general generators on task SR at Level~2, i.e. 2.4\% vs.\ SeedDance~2.0 and Wan~2.7 21.4\%.
Robot-specific models are sensitive to camera viewpoint and task domain, which limit generalization despite strong geometric precision.

\paragraph{In-domain fine-tuning improves appearance, not physics.}
Fine-tuning Wan~2.2 on in-domain episodes shifts the generated video appearance toward robotic motion and improves trajectory similarity, but does not improve task success rates significantly.
This suggests that injecting physical knowledge through robot video fine-tuning alone is insufficient, where the model learns the visual style of robot manipulation without acquiring the underlying physical constraints that drive task success.

\paragraph{Failure modes.}
Figure~\ref{fig:failures} illustrates three recurring failure categories: object levitation, phantom grasp, and kinematic breakdown.
Phantom grasps and kinematic breakdowns account for the majority of failed trials across all models.

\section{Conclusion}
\label{sec:conclusion}

The rapid progress of video generation has fueled excitement about using these models as world models and behavioral priors for robotics.
\ours\ puts this idea to a direct test: can the manipulation videos these models dream be grounded back into the physical world through robotic execution?
Evaluating 8 models across 101 tasks, we find the answer is a qualified yes.
Generative priors trained on internet-scale data encode physically meaningful motion, and several models achieve measurable execution success without any robot-specific supervision.
Yet visual quality remains a poor predictor of executability, and long-horizon tasks expose the limits of current models.
We hope \ours\ offers the community both the diagnostic tools and the motivation to close this gap.

{\small
\bibliography{references}
\bibliographystyle{icml2026_fogen}
}

\newpage
\appendix
\onecolumn

\section{VLM Visual Evaluation Details}
\label{app:vlm-eval}

Visual quality is assessed along three dimensions, each implemented as a separate VLM query run independently with Gemini 3 Pro and Qwen3-VL-Plus.
We describe the protocol and give a concrete prompt example for each dimension below.

\subsection*{Robot-Subject Stability}

Following the previous VLM evaluation work~\cite{deng2026rethinking}, two frames are sampled from each video: the first frame and the frame at 75\% of the total duration.
They are concatenated horizontally to form a side-by-side image, with the first frame on the left serving as a reference and the later frame on the right representing the generated output.
The VLM receives this composite image and is asked two questions in sequence.

\textit{Question 1} evaluates the robot subject (e.g., the robotic gripper or arm):

\begin{quote}
\small
The provided image shows two sequential frames from an AI-generated video about a robot doing a task.
The left frame is the correct reference image, while the right frame is the AI-generated video frame.
Focus on how `\textit{[robot subject]}' appears in both frames, and evaluate its consistency between the reference and the generated frame.

\noindent Note:\\
1) Pay special attention to distinguishing between robotic gripper and robotic hand. A robotic gripper usually has a small number of rigid gripping jaws or prongs, while a robotic hand has multiple articulated fingers.\\
2) Changes in orientation or position are acceptable and should not affect the consistency rating.\\
3) Do NOT assign option A or B lightly.

\noindent Question:\\
A: `\textit{[robot subject]}' in the right frame is clear and consistent with the left image.\\
B: mostly consistent, with minor visual issues.\\
C: noticeable inconsistencies in shape or structure.\\
D: highly inconsistent; transforms into another type of `\textit{[robot subject]}'.\\
E: \textit{[type-specific disappearance or substitution option]}.\\
Select the most suitable option and respond in JSON: \texttt{\{"option": "A", "explanation": "...", "adjust": "A"\}}.
\end{quote}

\textit{Question 2} repeats the same protocol for the manipulated object, with option E defined as the object being absent in the right frame.

The adjusted option from each question is combined into a pair, which is mapped to a score from 1 to 15 via a fixed lookup table.
The mapping is symmetric: (A, A) $\mapsto$ 15 and (E, E) $\mapsto$ 1, with intermediate combinations interpolated monotonically.

\subsection*{Physical Plausibility}

Six frames are sampled uniformly from the video and arranged into a $3 \times 2$ grid image.
The VLM receives this grid together with the camera viewpoint description and the task description, and is asked:

\begin{quote}
\small
The provided image presents sequential frames, arranged in a grid, from a \textit{[view]} perspective AI-generated task video about \textit{[task description]}.
Does this video comply with common-sense expectations for human-level interactions?

\noindent \textbf{A. Anomaly Checks:}\\
1) Physical grounding violation: any part of the robot appears floating, or intersecting/penetrating other geometry.\\
2) Spontaneous object appearance: any object that suddenly appears between frames without a plausible cause.\\
3) Non-contact attachment / false grasp: if the video involves grasping, check whether the object moves with the gripper without clear physical contact or closure.\\
If any anomaly is present, assign a low score (1--2).

\noindent \textbf{B. Common-Sense Consistency:}\\
Rate the video on a scale from 1 to 5, where 5 means fully consistent with human common sense and 1 means major violations.
Be cautious when assigning 4 or 5; do not give high scores lightly.
Respond in JSON: \texttt{\{"reason": "...", "score": 3\}}.
\end{quote}

\subsection*{Task Adherence}

The same $3 \times 2$ frame grid is used.
The VLM is asked:

\begin{quote}
\small
The provided image presents sequential frames, arranged in a grid, from a \textit{[view]} perspective AI-generated task video.
In this AI-generated video, does the robot successfully perform the task: ``\textit{[task description]}''?
Please rate the video on a scale from 1 to 5, where 5 indicates a perfect match and 1 indicates no relevance.
Be cautious when assigning scores of 4 or 5.
Respond in JSON: \texttt{\{"reason": "...", "score": 3\}}.
\end{quote}

\section{Model Details}
\label{app:models}

Table~\ref{tab:models} summarizes the evaluated models and their generation settings.
All models are conditioned on the initial scene image paired with a task instruction.
Resolution and video duration were set to the suitable option that does not exceed the task horizon: a short clip for Level~1 and Level~2 tasks, and a longer video for Level~3 multi-stage tasks.

\begin{table}[h]
  \caption{Models evaluated in \ours, organized by category.
    Open-weight models are marked with \cmark; closed-source API
    models are marked with \xmark.}
  \label{tab:models}
  \centering
  \small
  \begin{tabular}{llccccc}
    \toprule
    \textbf{Model} & \textbf{Category} & \textbf{Open} &
    \textbf{Resolution} & \textbf{Frames} & \textbf{FPS} & \textbf{Duration (s)} \\
    \midrule
    
    Hailuo 2.3~\cite{hailuo}           & Closed-source  & \xmark & $960{\times}960$   & $121 / 193$ & $24$ & $5 / 8$ \\
    Kling 3.0~\cite{kling}           & Closed-source  & \xmark & $960{\times}960$   & $121 / 193$ & $24$ & $5 / 8$ \\
    Wan~2.7~\cite{wan}           & Closed-source  & \xmark & $1440{\times}1440$ & $150 / 240$ & $30$ & $5 / 8$ \\
    SeedDance 2.0~\cite{seeddance}   & Closed-source  & \xmark & $960{\times}960$   & $121 / 193$ & $24$ & $5 / 8$ \\
    Veo~3.1~\cite{veo}           & Closed-source  & \xmark & $1280{\times}720$  & $144 / 192$ & $24$ & $6 / 8$ \\
    \midrule
    Wan~2.2~\cite{wan}           & Open-source    & \cmark & $480{\times}480$   & $81 / 129$  & $16$ & $5 / 8$ \\
    LTX 2.3~\cite{ltx}         & Open-source    & \cmark & $512{\times}512$   & $121$       & $24$ & $5$ \\
    \midrule
    Cosmos~Policy (rollout)~\cite{cosmos}  & Robot-specific & \cmark & $512{\times}512$   & $400$       & $15$ & $27$ \\
    Cosmos~Policy (generated)~\cite{cosmos} & Robot-specific & \cmark & $224{\times}224$   & $50$        & $1$  & $50$ \\
    \bottomrule
  \end{tabular}
\end{table}

\section{Video2Traj Implementation Details}
\label{app:v2t}

\begin{figure}[t]
  \centering
  \includegraphics[width=\linewidth]{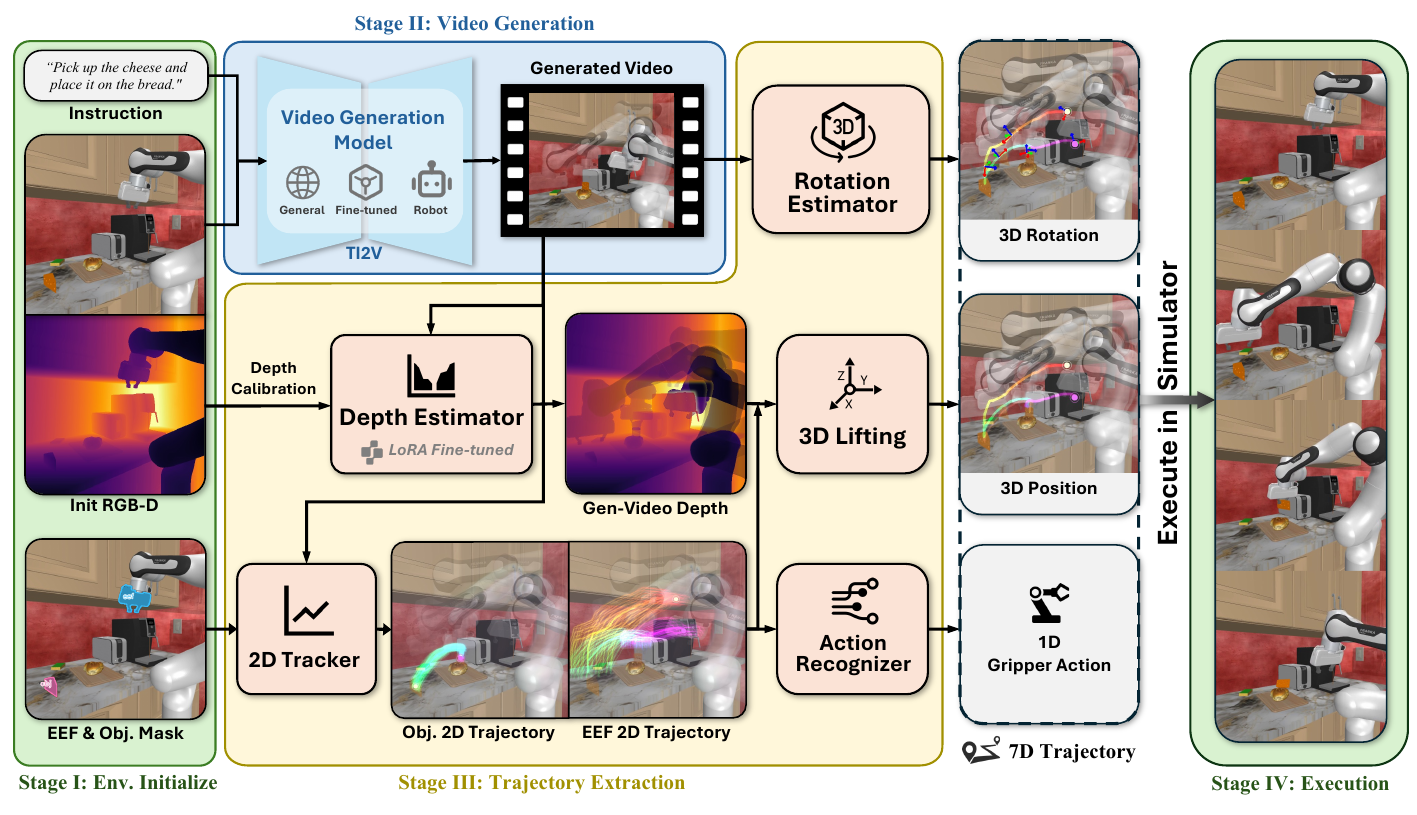}
  \caption{
  \textbf{Detailed video-to-execution pipeline.}
  The diagram expands the trajectory extraction and execution components of Figure~\ref{fig:pipeline}, showing how generated video is converted into 2D tracklets, calibrated depth, lifted 3D point trajectories, end-effector rotation, and gripper actions.
  These signals are fused into a 7D executable trajectory and replayed in the simulator for evaluation.
  }
  \label{fig:pipeline-details}
\end{figure}

Figure~\ref{fig:pipeline-details} provides an expanded view of the video-to-execution pipeline described in Section~\ref{sec:pipeline}.

\subsection{Trajectory Extraction and Execution Details}
\label{app:v2t-details}

\paragraph{2D point tracking.}

We initialize tracking regions on the first frame using simulator-provided segmentation masks when available, and fall back to visual region proposals such as manual boxes, open-vocabulary detection~\cite{gdino}, and SAM2~\cite{sam2} segmentation.
Within each region, query points are sampled from the mask using farthest-point sampling, using 3D sampling when initial depth and camera calibration are available and 2D sampling otherwise.
The selected points are tracked through the video using CoTracker~\cite{cotracker}, which returns per-frame point locations and visibility estimates.
During 3D lifting, points with low visibility, out-of-frame locations, or invalid depth are ignored on a per-frame basis.

\paragraph{Depth estimation.}
For generated videos, we use a robot-adapted DVD~\cite{zhang2026dvd} depth model to estimate temporally consistent video depth.
Specifically, we fine-tune low-rank adapters on top of the DVD DiT using robot rollout videos rendered from simulation, while keeping the remaining model weights frozen.
Each training sample consists of an RGB rollout clip paired with simulator-rendered metric depth.
Rather than supervising metric depth directly, we convert valid metric depth values into disparity, normalize them within each clip using robust percentiles, and train the LoRA adapters in this normalized-disparity space.
The training objective combines a masked reconstruction loss over valid depth pixels with spatial and temporal gradient matching losses, encouraging both sharp local depth boundaries and temporal consistency.
When end-effector weighting is used, simulator instance segmentation up-weights end-effector pixels in the reconstruction loss.
This end-effector emphasis improves the reliability of depth estimates around the robot hand, where small depth errors can lead to large errors in the recovered 3D end-effector trajectory.
At inference time, the LoRA weights are merged into the DVD DiT and used as the depth backend.
The predicted depth remains affine rather than metrically scaled; therefore, before 3D lifting, we align it to the first-frame simulator depth using a robust affine calibration over valid task regions.

In our implementation, the LoRA adapters are trained on fixed-length robot rollout clips at $512\times512$ resolution, with rank-512 adapters inserted into the DVD DiT attention and feed-forward projections.

\paragraph{3D Point Lifting}
Each valid tracked pixel is back-projected into the world frame using the calibrated depth and the simulator camera intrinsics and extrinsics.
The lifted point tracks are then summarized into per-frame visual-center trajectories for the end-effector and the manipulated object.
For each frame, the visual center is estimated robustly from valid lifted points, with carry-forward and interpolation used when too few reliable observations are available.
These visual-center trajectories provide the positional inputs for controller reference calibration and gripper schedule inference.

\paragraph{TCP Trajectory Extraction.}

The end-effector visual-center trajectory does not directly correspond to the robot tool-center point (TCP) or controller reference site.
We convert the recovered visual-center trajectory to a TCP trajectory through translation-only alignment to the simulator initial TCP pose: the offset is estimated from the median of the first valid visual-center observations and then applied to all frames.
When the controller reference site differs from the TCP, we further apply the fixed TCP-to-controller offset specified by the scene metadata.
This calibration uses the initial simulation state and does not require matching a full ground-truth rollout.

\paragraph{3D Rotation Estimation.}
In addition to the positional trajectory, we estimate the end-effector orientation for 6-DoF action generation.
Frame-wise rotations are estimated by rigidly aligning lifted 3D end-effector points to their first-frame anchors with Kabsch alignment.
The simulator-provided controller reference pose defines the initial geometric anchor for this alignment.
A lightweight temporal guard is applied to suppress implausible frame-to-frame rotation jumps while preserving the current translational trajectory.
The resulting orientation sequence is fused with the calibrated end-effector position trajectory before action generation.

\paragraph{Gripper Action Recognition.}
The gripper open-and-close schedule is inferred from the relative motion between the end-effector and the manipulated object.
We detect grasp and release events using geometric motion cues, including distance, relative velocity, co-motion, visibility, and invalid-track gating.
When task annotations are available, we identify the interaction mode of each manipulation stage and use its expected close/open pattern as a task prior, as summarized in Table~\ref{tab:gripper-interaction-priors}.
This prior constrains the number and ordering of gripper events, while their timing is still determined from motion evidence.
For multi-stage tasks, each stage is processed with its own target object and interaction prior, and the resulting stage-local schedules are merged into a single frame-aligned gripper schedule.
The resulting schedule is combined with the extracted end-effector trajectory during action generation.

\begin{table*}[t]
  \centering
  \caption{
  \textbf{Gripper action priors by interaction mode.}
  Each interaction mode specifies the expected number of close and open events used by the task-prior gripper recognizer.
  The occurrence count reports how often the interaction mode appears in the benchmark annotations; a single task may contain multiple interaction instances.
  }
  \label{tab:gripper-interaction-priors}
  \small
  \setlength{\tabcolsep}{5pt}
  \renewcommand{\arraystretch}{1.05}
  \begin{tabular}{@{}p{0.20\textwidth}p{0.47\textwidth}cccc@{}}
    \toprule
    \textbf{Interaction mode}
      & \textbf{Description}
      & \textbf{Level}
      & \textbf{Close}
      & \textbf{Open}
      & \textbf{\# Occ.} \\
    \midrule
    Pick-and-place
      & Grasp an object and place it at a target location or into another container.
      & level 2 \& 3  & 1 & 1 & 37 \\
    Object pushing
      & Move an object by pushing without requiring a grasp.
      & level 1 & 0 & 0 & 16 \\
    Drawer closing
      & Push a prismatic drawer from an open state toward a closed state.
      & level 1 & 0 & 0 & 14 \\
    Stacking
      & Grasp an object and place it on top of another object.
      & level 2 & 1 & 1 & 11 \\
    Object relocation
      & Grasp a removable object, lift it from its support or attachment, and place it at a target location.
      & level 2 & 1 & 1 & 10 \\
    Lever turning
      & Rotate an articulated lever or handle without a grasp-release cycle.
      & level 1 & 0 & 0 & 10 \\
    Door closing
      & Close a revolute articulated door.
      & level 1 & 0 & 0 & 7 \\
    Drawer opening
      & Pull or slide a prismatic drawer from a closed state toward an open state.
      & level 1 & 0 & 0 & 3 \\
    Switch activation
      & Activate an appliance or switch through direct contact.
      & level 1 & 0 & 0 & 1 \\
    \bottomrule
  \end{tabular}
\end{table*}

\paragraph{Closed-loop execution.}
The extracted trajectory is converted into a sequence of delta 6-DoF actions and gripper commands, and replayed in the robosuite~\cite{robosuite} operational-space controller.
Before each trial, the simulator scene is restored to the recorded initial state.
During action-mode execution, each retained trajectory checkpoint is reached through one or more controller steps.
At checkpoint boundaries, the executor compares the current configured controller-reference pose with the target checkpoint pose and applies a bounded number of corrective actions until the position and orientation errors fall below the configured tolerances.
In the current execution configuration, these tolerances are 5\,mm and 0.03\,rad.
This checkpoint-level correction reduces accumulated open-loop tracking error during simulation execution.

\subsection{Rollout-Video Reference Bounds}
\label{app:rollout-bounds}

To isolate the limitations of the video-to-execution pipeline from the limitations of generated videos, we also evaluate two rollout-video reference settings.
In both settings, the input video is the ground-truth rollout rendered from the simulator, rather than a generated video.
These rows therefore provide reference bounds on how well our video-to-execution pipeline can extract executable robot behavior when the visual motion is correct.

The \textit{Rollout Video} setting runs the same pipeline used for generated videos, including the learned depth estimator.
In contrast, \textit{Rollout Video w/ GT Depth} replaces the estimated depth with simulator-rendered metric depth while keeping the rest of the pipeline unchanged.
The gap between these two rows diagnoses the effect of depth estimation on downstream 3D lifting and execution.
As shown in Table~\ref{bench-exec-outcomes}, using ground-truth depth substantially improves task-level success, indicating that depth remains a major bottleneck in the current video-to-execution pipeline.
This is expected that small temporally inconsistent depth errors around the end-effector or manipulated object can be amplified after 3D lifting, leading to inaccurate TCP positions, contact timing errors, and ultimately lower execution success.

\section{Additional Quantitative Results}
\label{app:quant}

\subsection{Additional Visual Quality Scores under Different Instruction Settings}
\label{app:raw-prompt}

Table~\ref{tab:vq-judge-instruction-settings-by-level} provides a by-level breakdown of visual-quality scores across instruction settings, together with per-judge and judge-averaged scores.

\begin{table*}[t]
  \centering
  \caption{
    \textbf{By-level visual quality evaluation under different instruction settings.}
    Results are grouped by VLM judge, difficulty level, and instruction setting.
    Stab., Phys., and Task Adh. denote robot stability, physical plausibility, and task adherence.
    Judge Avg. reports the average of Gemini 3 Pro and Qwen3-VL-Plus scores. Higher is better ($\uparrow$).
    Rankings are computed separately within each instruction setting. \ranknote{}
  }
  \label{tab:vq-judge-instruction-settings-by-level}
  \tiny
  \setlength{\tabcolsep}{1.15pt}
  \renewcommand{\arraystretch}{0.92}
  \resizebox{\textwidth}{!}{%
  \begin{tabular}{@{}l*{27}{c}@{}}
    \toprule
    \textbf{Model}
      & \multicolumn{9}{c}{\textbf{Gemini 3 Pro}}
      & \multicolumn{9}{c}{\textbf{Qwen3-VL-Plus}}
      & \multicolumn{9}{c}{\textbf{Judge Avg.}} \\
    \cmidrule(lr){2-10}\cmidrule(lr){11-19}\cmidrule(lr){20-28}
      & \multicolumn{3}{c}{\textbf{Level~1}}
      & \multicolumn{3}{c}{\textbf{Level~2}}
      & \multicolumn{3}{c}{\textbf{Level~3}}
      & \multicolumn{3}{c}{\textbf{Level~1}}
      & \multicolumn{3}{c}{\textbf{Level~2}}
      & \multicolumn{3}{c}{\textbf{Level~3}}
      & \multicolumn{3}{c}{\textbf{Level~1}}
      & \multicolumn{3}{c}{\textbf{Level~2}}
      & \multicolumn{3}{c}{\textbf{Level~3}} \\
    \cmidrule(lr){2-4}\cmidrule(lr){5-7}\cmidrule(lr){8-10}
    \cmidrule(lr){11-13}\cmidrule(lr){14-16}\cmidrule(lr){17-19}
    \cmidrule(lr){20-22}\cmidrule(lr){23-25}\cmidrule(lr){26-28}
      & \textbf{Stab.} & \textbf{Phys.} & \textbf{Task}
      & \textbf{Stab.} & \textbf{Phys.} & \textbf{Task}
      & \textbf{Stab.} & \textbf{Phys.} & \textbf{Task}
      & \textbf{Stab.} & \textbf{Phys.} & \textbf{Task}
      & \textbf{Stab.} & \textbf{Phys.} & \textbf{Task}
      & \textbf{Stab.} & \textbf{Phys.} & \textbf{Task}
      & \textbf{Stab.} & \textbf{Phys.} & \textbf{Task}
      & \textbf{Stab.} & \textbf{Phys.} & \textbf{Task}
      & \textbf{Stab.} & \textbf{Phys.} & \textbf{Task} \\
    \midrule
    \multicolumn{28}{@{}l}{\scriptsize\textit{Videos generated from standard instructions; evaluated with standard instructions}} \\
    Hailuo~2.3 & 5.717 & 1.000 & 2.019 & 5.690 & 1.143 & 3.429 & 5.500 & 1.000 & 3.000 & \best{6.302} & 2.189 & \second{3.189} & 6.024 & 2.119 & \third{3.071} & 6.000 & \second{2.500} & 2.333 & 6.009 & 1.594 & 2.604 & 5.857 & 1.631 & 3.250 & 5.750 & 1.750 & 2.667 \\
    Kling~3.0 & \second{7.136} & \best{1.321} & \best{3.057} & 5.929 & 1.095 & 3.381 & 9.333 & 1.167 & \third{3.167} & 6.039 & 2.283 & \best{3.189} & 6.098 & 2.143 & 2.714 & \third{6.000} & \third{2.333} & 2.500 & \second{6.519} & \second{1.802} & \best{3.123} & 6.083 & 1.619 & 3.048 & 7.667 & 1.750 & 2.833 \\
    SeedDance~2.0 & 5.755 & \third{1.125} & 2.660 & 7.690 & \second{1.439} & \second{4.056} & 8.500 & \best{1.833} & \best{5.000} & 5.981 & 2.302 & \third{3.132} & 6.000 & \third{2.214} & 2.929 & 5.833 & 2.000 & 2.500 & 5.868 & \third{1.755} & \second{2.896} & 6.845 & \second{1.833} & \third{3.369} & 7.167 & \best{1.917} & \best{3.500} \\
    Veo~3.1 & 5.925 & 1.094 & \second{2.811} & 4.286 & \best{1.548} & \best{4.071} & 2.500 & \second{1.667} & \second{4.167} & 5.962 & 2.302 & 2.849 & 6.143 & \best{2.220} & 2.690 & 6.000 & 2.000 & \third{2.667} & 5.943 & 1.698 & 2.830 & 5.214 & \best{1.869} & \second{3.381} & 4.250 & \second{1.833} & \second{3.417} \\
    Wan~2.2 & 6.170 & 1.000 & 1.078 & 7.095 & 1.114 & 1.219 & \third{11.667} & 1.000 & 1.000 & 5.906 & \third{2.415} & 2.849 & 5.951 & 2.214 & 2.714 & 6.000 & 2.000 & 2.500 & 6.038 & 1.717 & 2.009 & 6.476 & \third{1.786} & 2.095 & \third{8.833} & 1.500 & 1.750 \\
    Wan~2.7 & 6.792 & \second{1.321} & \third{2.811} & \third{7.895} & \third{1.214} & \third{4.048} & 5.000 & \third{1.667} & 2.833 & \third{6.094} & 2.189 & 2.906 & 6.000 & 2.000 & \second{3.262} & 6.000 & 2.000 & \best{3.333} & 6.406 & 1.755 & \third{2.858} & 6.905 & 1.607 & \best{3.655} & 5.417 & \third{1.833} & \third{3.083} \\
    LTX~2.3 & 4.800 & 1.000 & 1.769 & 5.889 & 1.000 & 1.061 & 2.750 & 1.000 & 1.000 & 6.000 & \best{2.453} & 2.788 & 6.000 & \second{2.214} & 2.929 & 6.000 & 2.000 & 2.250 & 5.786 & \best{2.123} & 2.547 & 5.964 & 1.774 & 2.179 & 4.375 & 1.500 & 1.625 \\
    Wan 2.2-LoRA$_{2K}$ & 6.132 & 1.075 & 1.226 & \second{8.381} & 1.024 & 1.190 & 8.000 & 1.000 & 1.167 & \second{6.208} & 2.208 & 2.895 & 6.171 & 2.119 & \best{3.270} & \best{7.000} & \best{2.667} & 2.333 & 6.170 & 1.642 & 1.792 & \second{7.238} & 1.571 & 2.119 & 7.500 & 1.833 & 1.750 \\
    Wan 2.2-LoRA$_{7K}$ & 6.667 & 1.020 & 1.180 & 7.821 & 1.000 & 1.167 & \second{12.167} & 1.000 & 1.000 & 6.000 & 2.231 & 3.000 & \best{6.524} & 2.167 & 2.878 & 6.000 & 2.167 & 2.500 & 6.144 & 1.651 & 2.132 & \third{7.107} & 1.583 & 2.000 & \second{9.083} & 1.583 & 1.750 \\
    CosmosPolicy-DefaultCam & \third{6.800} & 1.000 & 1.595 & 7.750 & 1.158 & 1.000 & 11.167 & 1.000 & 1.000 & 6.073 & \second{2.439} & 3.119 & \third{6.316} & 2.053 & 3.053 & \second{6.000} & 2.000 & \second{2.833} & \third{6.427} & 1.702 & 2.357 & 7.026 & 1.605 & 2.026 & \best{9.083} & 1.500 & 1.917 \\
    CosmosPolicy-BenchCam & \best{8.000} & 1.000 & 1.357 & \best{9.706} & 1.000 & 1.000 & \best{12.500} & 1.000 & 1.000 & 5.810 & 2.341 & 2.927 & \second{6.316} & 2.211 & 2.789 & 4.833 & 2.000 & 2.333 & \best{6.881} & 1.655 & 2.119 & \best{7.816} & 1.605 & 1.895 & 8.667 & 1.500 & 1.667 \\
    \midrule
    \multicolumn{28}{@{}l}{\scriptsize\textit{Videos generated from enhanced instructions; evaluated with standard instructions}} \\
    Hailuo~2.3 & 4.981 & 1.094 & 2.302 & 5.381 & 1.167 & 3.310 & 3.333 & 1.000 & 1.833 & 5.941 & 2.340 & 2.849 & 5.881 & \best{2.310} & \third{2.976} & 5.500 & 2.000 & \best{3.000} & 5.415 & 1.717 & 2.575 & 5.631 & 1.738 & 3.143 & 4.417 & 1.500 & 2.417 \\
    Kling~3.0 & \second{7.333} & \best{1.300} & \third{2.938} & 6.316 & 1.286 & \best{4.048} & 7.000 & \best{1.667} & 3.000 & 5.887 & \second{2.377} & \second{3.075} & \third{6.220} & 2.190 & 2.833 & \second{6.000} & 1.833 & 1.667 & \third{6.585} & \best{1.877} & \third{3.019} & 6.321 & \third{1.738} & \second{3.440} & 6.500 & \best{1.750} & 2.333 \\
    SeedDance~2.0 & 6.245 & 1.189 & \second{3.321} & 7.786 & 1.310 & 3.976 & \second{11.000} & \second{1.167} & \best{4.000} & 6.038 & 2.321 & 2.925 & 5.976 & \third{2.214} & 2.738 & 6.000 & 2.000 & 2.667 & 6.142 & 1.755 & \second{3.123} & 6.881 & \second{1.762} & \third{3.357} & \second{8.500} & \third{1.583} & \best{3.333} \\
    Veo~3.1 & 6.389 & 1.056 & \best{3.444} & 4.909 & \third{1.364} & \second{4.000} & 5.167 & \third{1.167} & \third{3.667} & 6.000 & 2.333 & \best{3.500} & 5.909 & 1.818 & \second{3.000} & 6.000 & 2.000 & \second{2.833} & 6.194 & 1.694 & \best{3.472} & 5.409 & 1.591 & \best{3.500} & 5.583 & 1.583 & \second{3.250} \\
    Wan~2.2 & 5.434 & 1.075 & 2.094 & 5.357 & 1.119 & 2.833 & 6.000 & 1.000 & 1.167 & 6.057 & \third{2.358} & 2.774 & 6.095 & 2.095 & 2.714 & 6.000 & 2.000 & 2.500 & 5.745 & 1.717 & 2.434 & 5.726 & 1.607 & 2.774 & 6.000 & 1.500 & 1.833 \\
    Wan~2.7 & \third{7.321} & \second{1.226} & 2.925 & \third{8.143} & \best{1.762} & \third{4.000} & 6.500 & 1.167 & \second{3.833} & 6.000 & 2.340 & \third{3.000} & 5.976 & 2.119 & 2.643 & 6.000 & \best{2.167} & 2.500 & \second{6.660} & 1.783 & 2.962 & \third{7.060} & \best{1.940} & 3.321 & 6.250 & \second{1.667} & \third{3.167} \\
    LTX~2.3 & 4.887 & 1.000 & 2.600 & 3.900 & 1.000 & 1.852 & 3.833 & 1.000 & 1.333 & \second{6.321} & 2.151 & 2.962 & 6.000 & 2.100 & 2.641 & \third{6.000} & 2.000 & 2.167 & 5.321 & \second{1.821} & 2.868 & 4.950 & 1.575 & 2.282 & 4.917 & 1.500 & 1.750 \\
    Wan 2.2-LoRA$_{2K}$ & 6.038 & 1.132 & 2.057 & 7.071 & \second{1.405} & 2.929 & 4.667 & 1.000 & 1.167 & 5.952 & 2.216 & 3.000 & 6.000 & 2.065 & 2.825 & 6.000 & 2.000 & 2.500 & 5.962 & 1.651 & 2.396 & 6.905 & 1.702 & 2.845 & 5.333 & 1.500 & 1.833 \\
    Wan 2.2-LoRA$_{7K}$ & 6.269 & \third{1.208} & 2.216 & 7.128 & 1.071 & 2.476 & 4.500 & 1.000 & 1.000 & \third{6.192} & \best{2.442} & 2.980 & 6.073 & 2.095 & 2.833 & 6.000 & 2.000 & \third{2.833} & 6.231 & \third{1.811} & 2.594 & 6.595 & 1.583 & 2.655 & 5.250 & 1.500 & 1.917 \\
    CosmosPolicy-DefaultCam & 6.610 & 1.000 & 1.381 & \best{9.312} & 1.053 & 1.000 & \third{9.500} & 1.000 & 1.000 & 6.073 & 2.333 & 2.929 & \best{6.526} & \second{2.263} & 2.737 & 5.500 & \third{2.000} & 2.500 & 6.321 & 1.667 & 2.155 & \best{7.816} & 1.658 & 1.868 & \third{7.500} & 1.417 & 1.750 \\
    CosmosPolicy-BenchCam & \best{8.590} & 1.000 & 1.119 & \second{8.765} & 1.000 & 1.000 & \best{11.833} & 1.000 & 1.000 & \best{6.439} & 2.250 & 2.976 & \second{6.316} & 1.947 & \best{3.222} & \best{6.000} & \second{2.000} & 2.167 & \best{7.405} & 1.595 & 2.048 & \second{7.447} & 1.474 & 2.053 & \best{8.917} & 1.500 & 1.583 \\
    \midrule
    \multicolumn{28}{@{}l}{\scriptsize\textit{Videos generated from enhanced instructions; evaluated with enhanced instructions}} \\
    Hailuo~2.3 & 5.094 & 1.019 & 1.415 & 5.405 & 1.000 & 1.643 & 5.333 & 1.000 & 1.000 & 6.294 & \second{4.731} & 2.808 & 5.976 & 4.452 & \second{2.905} & 6.000 & \third{4.667} & 2.667 & 5.698 & 2.840 & 2.094 & 5.690 & 2.726 & 2.274 & 5.667 & 2.833 & 1.833 \\
    Kling~3.0 & 6.604 & 1.019 & \best{1.755} & 6.049 & 1.049 & \best{2.024} & \second{12.000} & \third{1.000} & 1.333 & 6.058 & \best{4.769} & 2.642 & 6.095 & 4.548 & 2.833 & 6.000 & 4.500 & 2.667 & 6.387 & 2.858 & \second{2.198} & 6.071 & 2.845 & \second{2.429} & 6.500 & \best{3.583} & 2.000 \\
    SeedDance~2.0 & 6.660 & 1.000 & \third{1.566} & \third{8.143} & \third{1.119} & \third{1.857} & \third{10.167} & 1.000 & \best{2.000} & 6.096 & \third{4.717} & 2.830 & 6.024 & 4.548 & \third{2.905} & 6.000 & 4.667 & \second{3.000} & 6.396 & 2.858 & \third{2.198} & \third{7.083} & 2.833 & \third{2.381} & \second{8.083} & 2.833 & \second{2.500} \\
    Veo~3.1 & \second{6.889} & \third{1.056} & 1.444 & 5.273 & \best{1.364} & \second{2.000} & 4.667 & 1.000 & \second{2.000} & \best{6.588} & 4.389 & 2.778 & 5.455 & \best{4.909} & \best{3.091} & 5.500 & 4.333 & \best{3.167} & \second{6.667} & 2.722 & 2.111 & 5.364 & \second{3.136} & \best{2.545} & 5.083 & 2.667 & \best{2.583} \\
    Wan~2.2 & 5.358 & \best{1.094} & 1.396 & 4.929 & 1.000 & 1.452 & 4.000 & 1.000 & 1.000 & 6.075 & 4.660 & \third{2.885} & \second{6.476} & 4.619 & 2.833 & 6.000 & 4.667 & 2.667 & 5.717 & \second{2.877} & 2.132 & 5.702 & 2.810 & 2.143 & 5.000 & 2.833 & 1.833 \\
    Wan~2.7 & 6.434 & 1.038 & \second{1.604} & 8.048 & \second{1.143} & 1.738 & 6.333 & 1.000 & \third{1.833} & 5.962 & 4.585 & 2.774 & 6.000 & 4.571 & 2.810 & \best{7.000} & 4.667 & 2.667 & 6.217 & 2.811 & 2.189 & 7.024 & \third{2.857} & 2.274 & 6.667 & 2.833 & \third{2.250} \\
    LTX~2.3 & 4.571 & 1.000 & 1.208 & 4.500 & 1.000 & 1.115 & 3.667 & 1.000 & 1.000 & \second{6.571} & 4.642 & \best{2.942} & 6.025 & \third{4.718} & 2.650 & 6.000 & 4.167 & 2.667 & 6.107 & \best{3.726} & \best{2.491} & 5.875 & \best{3.487} & 2.175 & 4.833 & 2.583 & 1.833 \\
    Wan 2.2-LoRA$_{2K}$ & 5.852 & \second{1.077} & 1.321 & 6.659 & 1.000 & 1.476 & 5.333 & 1.000 & 1.000 & 6.170 & 4.623 & \second{2.933} & 6.071 & 4.595 & 2.853 & \second{6.167} & 4.333 & 2.500 & 6.142 & \third{2.877} & 1.981 & 6.429 & 2.798 & 2.048 & 5.750 & 2.667 & 1.750 \\
    Wan 2.2-LoRA$_{7K}$ & 5.846 & 1.038 & 1.314 & 7.000 & 1.000 & 1.476 & 6.833 & 1.000 & 1.000 & 6.135 & 4.596 & 2.808 & 5.927 & 4.405 & 2.738 & 6.000 & 4.667 & 2.833 & 6.106 & 2.821 & 2.066 & 6.488 & 2.738 & 2.107 & 6.417 & 2.833 & 1.917 \\
    CosmosPolicy-DefaultCam & \third{6.690} & 1.048 & 1.024 & \second{8.812} & 1.000 & 1.105 & 9.500 & \second{1.000} & 1.000 & \third{6.357} & 4.595 & 2.810 & \third{6.471} & \second{4.889} & 2.684 & 6.000 & \best{4.833} & 2.167 & \third{6.524} & 2.821 & 1.917 & \second{7.711} & 2.842 & 1.895 & \third{7.750} & \second{2.917} & 1.583 \\
    CosmosPolicy-BenchCam & \best{8.850} & 1.000 & 1.000 & \best{8.947} & 1.000 & 1.000 & \best{12.167} & \best{1.000} & 1.000 & 6.098 & 4.619 & 2.854 & \best{6.556} & 4.556 & 2.632 & \third{6.000} & \second{4.667} & \third{2.833} & \best{7.500} & 2.810 & 1.905 & \best{7.842} & 2.684 & 1.816 & \best{9.083} & \third{2.833} & 1.917 \\
    \bottomrule
  \end{tabular}
  }
\end{table*}

\subsection{Trajectory Similarity Metrics under Different Instruction Settings}

Table~\ref{tab:traj-union-instruction-breakdown} reports trajectory similarity scores
under standard and enhanced instructions, complementing Table~\ref{tab:traj-metrics-union-all-models} in the main paper.

\begin{table*}[t]
  \caption{
  \textbf{Trajectory similarity results under different instruction settings.}
  Results are reported for videos generated from enhanced and standard instructions.
  EEF vis, EEF tcp, and OBJ denote the end-effector visual center, end-effector tool center point, and manipulated object.
  HSD, DYN, and NDTW measure trajectory shape, dynamics, and temporal-alignment similarity.
  Higher is better ($\uparrow$). \ranknote{}
  }
  \label{tab:traj-union-instruction-breakdown}
  \centering
  \small
  \setlength{\tabcolsep}{3.5pt}
  \renewcommand{\arraystretch}{0.96}
  \begin{tabular}{@{}lccccccccc@{}}
    \toprule
    \textbf{Model}
    & \multicolumn{3}{c}{\textbf{EEF vis}}
    & \multicolumn{3}{c}{\textbf{EEF tcp}}
    & \multicolumn{3}{c}{\textbf{OBJ}} \\
    \cmidrule(lr){2-4}\cmidrule(lr){5-7}\cmidrule(lr){8-10}
    & \textbf{HSD} & \textbf{DYN} & \textbf{NDTW}
    & \textbf{HSD} & \textbf{DYN} & \textbf{NDTW}
    & \textbf{HSD} & \textbf{DYN} & \textbf{NDTW} \\
    \midrule
    \multicolumn{10}{@{}l}{\scriptsize\textit{Videos generated from enhanced instructions}} \\
    Hailuo~2.3            & 0.636 & 0.654 & 0.721 & 0.725 & 0.722 & \third{0.838} & 0.536 & \third{0.814} & 0.699 \\
    Kling~3.0             & \third{0.758} & \third{0.761} & \second{0.844} & \third{0.747} & 0.755 & 0.834 & 0.514 & 0.764 & 0.678 \\
    SeedDance~2.0         & 0.679 & 0.671 & 0.785 & 0.709 & 0.706 & 0.827 & 0.490 & 0.758 & 0.646 \\
    Veo~3.1               & 0.601 & 0.642 & 0.671 & 0.714 & 0.768 & 0.808 & 0.511 & 0.774 & 0.662 \\
    Wan~2.2               & 0.625 & 0.665 & 0.736 & 0.657 & 0.706 & 0.785 & 0.427 & 0.645 & 0.536 \\
    Wan~2.7               & \second{0.777} & \second{0.807} & \best{0.861} & \second{0.780} & \second{0.814} & \best{0.870} & \second{0.582} & \second{0.847} & \third{0.736} \\
    LTX~2.3               & 0.551 & 0.595 & 0.606 & 0.706 & \third{0.778} & 0.797 & \third{0.573} & 0.811 & \second{0.737} \\
    Wan 2.2-LoRA$_{2K}$   & 0.634 & 0.669 & 0.745 & 0.663 & 0.698 & 0.783 & 0.427 & 0.623 & 0.535 \\
    Wan 2.2-LoRA$_{7K}$   & 0.666 & 0.716 & 0.781 & 0.669 & 0.723 & 0.791 & 0.397 & 0.637 & 0.478 \\
    CosmosPolicy-BenchCam & \best{0.785} & \best{0.845} & \third{0.831} & \best{0.791} & \best{0.866} & \second{0.848} & \best{0.637} & \best{0.898} & \best{0.806} \\
    \midrule
    \multicolumn{10}{@{}l}{\scriptsize\textit{Videos generated from standard instructions}} \\
    Hailuo~2.3            & 0.609 & 0.622 & 0.688 & 0.707 & 0.708 & 0.805 & 0.574 & 0.827 & 0.740 \\
    Kling~3.0             & \third{0.708} & 0.718 & \best{0.819} & \third{0.720} & 0.724 & \second{0.838} & 0.520 & 0.769 & 0.700 \\
    SeedDance~2.0         & 0.704 & 0.724 & 0.808 & 0.691 & 0.703 & 0.804 & 0.480 & 0.744 & 0.671 \\
    Veo~3.1               & 0.473 & 0.485 & 0.530 & 0.718 & 0.741 & 0.817 & 0.542 & 0.804 & 0.687 \\
    Wan~2.2               & 0.652 & 0.751 & 0.718 & 0.668 & 0.770 & 0.744 & 0.583 & 0.789 & 0.639 \\
    Wan~2.7               & \second{0.729} & 0.749 & \third{0.815} & \second{0.745} & 0.764 & \best{0.854} & \third{0.615} & \second{0.857} & \second{0.765} \\
    LTX~2.3               & 0.586 & 0.658 & 0.640 & 0.714 & \third{0.785} & 0.797 & \best{0.631} & \best{0.864} & \third{0.748} \\
    Wan 2.2-LoRA$_{2K}$   & 0.656 & \third{0.765} & 0.727 & 0.675 & 0.782 & 0.755 & 0.569 & 0.769 & 0.627 \\
    Wan 2.2-LoRA$_{7K}$   & 0.676 & \second{0.791} & 0.752 & 0.686 & \second{0.807} & 0.769 & 0.534 & 0.762 & 0.574 \\
    CosmosPolicy-BenchCam & \best{0.756} & \best{0.822} & \second{0.816} & \best{0.750} & \best{0.811} & \third{0.823} & \second{0.622} & \third{0.849} & \best{0.789} \\
    \bottomrule
  \end{tabular}
\end{table*}

\begin{table*}[t]
  \centering
  \caption{
    \textbf{Trajectory execution feasibility under different instruction settings.}
    Results are reported for trajectories extracted from videos generated with standard and enhanced instructions, broken down by task difficulty and overall.
    E-SR measures checkpoint reachability ($\uparrow$), while nDTW, Pos95/Rot95, and Smth measure TCP tracking disagreement, 95th-percentile position/rotation error, and executed-trajectory smoothness ($\downarrow$).
    The Overall block aggregates over all active tasks without stratifying by difficulty.
    \ranknote{}
    }
  \label{tab:exec_metrics_by_level_penalty_instruction_ablation}
  \scriptsize
  \setlength{\tabcolsep}{1.0pt}
  \renewcommand{\arraystretch}{0.94}
  \resizebox{\textwidth}{!}{%
  \begin{tabular}{@{}l*{20}{c}@{}}
    \toprule
    \textbf{Model}
      & \multicolumn{5}{c}{\textbf{Level 1}}
      & \multicolumn{5}{c}{\textbf{Level 2}}
      & \multicolumn{5}{c}{\textbf{Level 3}}
      & \multicolumn{5}{c}{\textbf{Overall}} \\
    \cmidrule(lr){2-6}\cmidrule(lr){7-11}\cmidrule(lr){12-16}\cmidrule(lr){17-21}
      & \textbf{E-SR}$\uparrow$ & \textbf{nDTW}$\downarrow$ & \textbf{Pos95}$\downarrow$ & \textbf{Rot95}$\downarrow$ & \textbf{Smth}$\downarrow$
      & \textbf{E-SR}$\uparrow$ & \textbf{nDTW}$\downarrow$ & \textbf{Pos95}$\downarrow$ & \textbf{Rot95}$\downarrow$ & \textbf{Smth}$\downarrow$
      & \textbf{E-SR}$\uparrow$ & \textbf{nDTW}$\downarrow$ & \textbf{Pos95}$\downarrow$ & \textbf{Rot95}$\downarrow$ & \textbf{Smth}$\downarrow$
      & \textbf{E-SR}$\uparrow$ & \textbf{nDTW}$\downarrow$ & \textbf{Pos95}$\downarrow$ & \textbf{Rot95}$\downarrow$ & \textbf{Smth}$\downarrow$ \\
    \midrule
    \multicolumn{21}{@{}l}{\scriptsize\textit{Videos generated from standard instructions}} \\
    Hailuo 2.3 & 0.506 & 37.227 & 81.936 & 24.791 & 17.171 & 0.523 & 21.510 & 91.097 & 24.127 & 19.415 & 0.625 & 5.238 & 2.585 & 13.067 & 19.575 & 0.520 & 28.791 & 81.031 & 23.788 & 18.247 \\
    Kling 3.0 & 0.426 & \third{7.367} & 8.625 & 21.881 & 16.864 & 0.503 & 21.960 & 260.739 & 25.128 & 17.416 & 0.595 & 4.307 & 1.932 & 12.289 & 19.976 & 0.468 & 13.253 & 113.067 & 22.601 & 17.279 \\
    SeedDance 2.0 & 0.447 & 9.973 & 11.458 & 25.139 & 16.516 & 0.552 & 10.896 & 63.382 & 15.558 & 20.508 & 0.650 & 3.706 & 4.487 & 10.136 & 18.970 & 0.503 & 9.985 & 32.636 & 20.309 & 18.322 \\
    Veo 3.1 & \second{0.537} & 46.636 & 181.877 & 21.522 & 16.294 & 0.470 & 9.510 & 8.082 & \third{12.709} & 23.351 & 0.694 & 6.118 & \best{1.074} & 12.964 & 18.478 & 0.519 & 28.790 & 98.865 & \third{17.407} & 19.359 \\
    Wan 2.2 & 0.438 & 9.324 & \third{8.078} & 22.117 & 14.511 & 0.513 & \third{7.882} & 6.318 & 22.284 & \second{16.336} & 0.604 & 3.296 & \third{1.697} & 11.199 & \second{16.647} & 0.479 & 8.366 & 6.967 & 21.502 & 15.397 \\
    Wan 2.7 & 0.496 & 9.046 & 9.063 & 23.789 & 17.561 & \second{0.614} & 71.339 & 266.288 & 18.331 & 16.997 & 0.647 & 6.018 & 6.664 & 16.571 & \third{16.886} & \third{0.554} & 34.770 & 115.885 & 21.120 & 17.287 \\
    LTX 2.3 & 0.407 & 11.584 & 12.131 & 22.198 & 15.167 & 0.399 & 13.181 & 10.059 & 25.779 & 19.875 & 0.427 & 3.077 & \second{1.301} & \third{3.628} & \best{14.120} & 0.404 & 11.918 & 10.814 & 22.901 & 17.122 \\
    Wan 2.2-LoRA$_{2K}$ & 0.430 & 8.196 & 8.283 & 20.251 & 14.223 & 0.513 & 7.982 & \third{6.007} & 19.361 & \best{14.935} & \third{0.710} & \best{2.449} & 1.849 & 5.533 & 18.514 & 0.481 & \third{7.766} & \third{6.954} & 18.969 & \second{14.774} \\
    Wan 2.2-LoRA$_{7K}$ & 0.448 & 9.192 & 9.472 & \third{17.756} & \third{13.814} & 0.468 & 9.072 & 7.162 & 28.689 & \third{16.424} & 0.656 & \second{2.540} & 2.212 & 6.083 & 17.035 & 0.469 & 8.743 & 8.066 & 21.507 & \third{15.103} \\
    CosmosPolicy-DefaultCam & \best{0.549} & \best{5.198} & \best{4.268} & \best{5.737} & \best{10.936} & \best{0.877} & \best{2.651} & \best{3.120} & \best{4.247} & 16.939 & \best{0.896} & \third{2.767} & 3.102 & \best{1.983} & 19.007 & \best{0.706} & \best{3.994} & \best{3.721} & \best{4.895} & \best{13.911} \\
    CosmosPolicy-BenchCam & \third{0.522} & \second{5.387} & \second{4.706} & \second{6.661} & \second{13.644} & \third{0.590} & \second{4.644} & \second{4.777} & \second{5.781} & 17.706 & \second{0.734} & 3.575 & 4.120 & \second{3.269} & 17.770 & \second{0.563} & \second{4.970} & \second{4.701} & \second{6.094} & 15.578 \\
    \midrule
    \multicolumn{21}{@{}l}{\scriptsize\textit{Videos generated from enhanced instructions}} \\
    Hailuo 2.3 & 0.510 & 15.268 & 24.257 & 26.524 & 16.560 & 0.496 & 481.331 & 1576.635 & 20.223 & 16.091 & \second{0.754} & 7.866 & \best{2.143} & 13.478 & \second{17.132} & 0.519 & 208.637 & 668.487 & 23.149 & \third{16.399} \\
    Kling 3.0 & 0.416 & 10.561 & 9.983 & 34.767 & 19.569 & 0.524 & 25.571 & 339.954 & 29.949 & 17.474 & \third{0.619} & \second{3.023} & \second{2.428} & 8.987 & 19.790 & 0.473 & 16.355 & 146.750 & 31.198 & 18.711 \\
    SeedDance 2.0 & 0.427 & 30.630 & 259.515 & 32.694 & 22.642 & \third{0.564} & 11.490 & 67.771 & 15.155 & 20.039 & 0.557 & 6.775 & 6.375 & 13.876 & 17.271 & 0.492 & 21.254 & 164.742 & 24.393 & 21.240 \\
    Veo 3.1 & 0.507 & 11.017 & 12.101 & 27.078 & 15.990 & 0.556 & 8.687 & 9.374 & \third{11.892} & 16.462 & 0.569 & 8.814 & 7.202 & \third{8.352} & \best{16.819} & 0.535 & 9.706 & 10.400 & 18.239 & \second{16.283} \\
    Wan 2.2 & 0.459 & 216.255 & 1828.324 & 19.910 & \second{13.620} & 0.431 & 8.588 & 9.005 & 23.806 & \best{14.853} & 0.464 & 4.859 & 2.540 & 13.000 & 86.990 & 0.448 & 117.340 & 963.313 & 21.061 & 18.491 \\
    Wan 2.7 & \third{0.530} & 8.885 & 9.050 & 25.006 & 20.850 & \second{0.620} & \third{6.968} & 16.354 & 17.807 & 18.041 & 0.585 & 5.018 & 6.442 & 13.730 & 20.551 & \third{0.571} & \third{7.858} & 11.933 & 21.377 & 19.664 \\
    LTX 2.3 & 0.437 & 7.994 & 7.503 & 19.398 & 15.343 & 0.385 & 9.448 & \third{8.943} & 16.516 & 21.649 & 0.077 & 29.953 & 5.108 & 43.539 & 109.881 & 0.397 & 9.708 & \third{7.968} & 19.550 & 22.740 \\
    Wan 2.2-LoRA$_{2K}$ & 0.500 & \third{7.218} & \third{7.126} & \third{15.732} & 14.271 & 0.416 & 9.637 & 28.208 & 19.303 & \third{15.203} & 0.514 & \third{3.867} & \third{2.463} & 10.322 & 87.312 & 0.466 & 8.025 & 15.616 & \third{16.845} & 18.998 \\
    Wan 2.2-LoRA$_{7K}$ & 0.494 & 8.585 & 7.522 & 20.149 & \best{13.521} & 0.422 & 8.845 & 10.052 & 25.842 & \second{15.010} & 0.450 & 5.091 & 2.909 & 13.539 & 90.370 & 0.461 & 8.484 & 8.308 & 22.069 & 18.757 \\
    CosmosPolicy-DefaultCam & \best{0.774} & \best{3.554} & \best{3.587} & \best{4.161} & \third{14.140} & \best{0.805} & \best{3.159} & \best{3.624} & \best{4.007} & 17.770 & \best{0.887} & \best{2.821} & 3.239 & \best{2.655} & 17.922 & \best{0.794} & \best{3.346} & \best{3.582} & \best{4.008} & \best{15.874} \\
    CosmosPolicy-BenchCam & \second{0.732} & \second{3.891} & \second{4.004} & \second{5.861} & 14.969 & 0.536 & \second{5.011} & \second{5.471} & \second{5.166} & 18.344 & 0.590 & 4.620 & 5.250 & \second{4.807} & \third{17.266} & \second{0.642} & \second{4.400} & \second{4.688} & \second{5.510} & 16.509 \\
    \bottomrule
  \end{tabular}
  }
  \vspace{2pt}
  \begin{minipage}{0.97\textwidth}
    \scriptsize
    \emph{Note.}
    Scores use the same trajectory-validity adjustment as Table~\ref{tab:exec_metrics_by_level} to penalize unrealistically short motions.
    All reported metrics are means over active UIDs within each difficulty level or over all active UIDs for Overall.
    Rankings are computed separately within each instruction setting.
  \end{minipage}
\end{table*}

\begin{table*}[t]
  \centering
  \caption{
  \textbf{Task-level execution results under different instruction settings.}
  Results are reported for videos generated from standard and enhanced instructions, broken down by task difficulty and overall.
  SR-B is the binary task success rate and SR-P is a continuous partial-completion score.
  Rel, Place, Art, and Core report sub-goal completion for end-effector release, target placement, articulation progress, and core sub-goal fraction, whose availability depends on the task category and difficulty.
  Higher is better for all metrics ($\uparrow$). \ranknote{}
  }
  \label{tab:bench-aggregate-instruction-ablation}
  \scriptsize
  \setlength{\tabcolsep}{2.0pt}
  \renewcommand{\arraystretch}{0.94}
  \resizebox{\textwidth}{!}{%
  \begin{tabular}{@{}l*{20}{c}@{}}
    \toprule
    \textbf{Model}
      & \multicolumn{3}{c}{\textbf{Level 1}}
      & \multicolumn{6}{c}{\textbf{Level 2}}
      & \multicolumn{5}{c}{\textbf{Level 3}}
      & \multicolumn{6}{c}{\textbf{Overall}} \\
    \cmidrule(lr){2-4} \cmidrule(lr){5-10} \cmidrule(lr){11-15} \cmidrule(lr){16-21}
      & \textbf{SR-B}$\uparrow$ & \textbf{SR-P}$\uparrow$ & \textbf{Art}$\uparrow$
      & \textbf{SR-B}$\uparrow$ & \textbf{SR-P}$\uparrow$ & \textbf{Rel}$\uparrow$ & \textbf{Place}$\uparrow$ & \textbf{Art}$\uparrow$ & \textbf{Core}$\uparrow$
      & \textbf{SR-B}$\uparrow$ & \textbf{SR-P}$\uparrow$ & \textbf{Rel}$\uparrow$ & \textbf{Place}$\uparrow$ & \textbf{Core}$\uparrow$
      & \textbf{SR-B}$\uparrow$ & \textbf{SR-P}$\uparrow$ & \textbf{Rel}$\uparrow$ & \textbf{Place}$\uparrow$ & \textbf{Art}$\uparrow$ & \textbf{Core}$\uparrow$ \\
    \midrule
    \multicolumn{21}{@{}l}{\scriptsize\textit{Videos generated from standard instructions}} \\
    Hailuo 2.3 & 0.075 & 0.179 & 0.140 & \second{0.167} & \third{0.610} & 0.778 & 0.297 & \third{0.774} & \third{0.172} & 0.000 & \third{0.312} & \second{0.625} & 0.000 & 0.000 & 0.107 & 0.365 & 0.754 & 0.238 & 0.262 & \third{0.138} \\
    Kling 3.0 & 0.132 & \third{0.256} & \best{0.245} & \second{0.167} & 0.583 & 0.507 & \best{0.466} & 0.718 & \best{0.359} & \best{0.125} & \second{0.344} & \third{0.500} & \best{0.188} & \best{0.188} & \best{0.146} & 0.396 & 0.506 & \best{0.410} & \best{0.336} & \best{0.325} \\
    SeedDance 2.0 & \third{0.151} & \second{0.283} & \second{0.221} & \second{0.167} & \second{0.656} & \third{0.834} & 0.293 & 0.756 & 0.156 & 0.000 & \best{0.406} & \best{0.750} & \second{0.062} & \second{0.062} & \best{0.146} & \best{0.445} & \second{0.821} & \third{0.247} & \second{0.324} & \third{0.138} \\
    Veo 3.1 & 0.038 & 0.126 & 0.088 & \third{0.119} & 0.586 & \second{0.844} & 0.276 & 0.701 & 0.141 & 0.000 & \second{0.344} & \second{0.625} & \second{0.062} & \second{0.062} & 0.068 & 0.331 & \third{0.809} & 0.233 & 0.206 & 0.125 \\
    Wan 2.2 & 0.038 & 0.110 & 0.057 & 0.024 & 0.472 & 0.541 & 0.276 & \best{0.809} & 0.141 & 0.000 & 0.188 & 0.375 & 0.000 & 0.000 & 0.029 & 0.264 & 0.514 & 0.221 & 0.202 & 0.113 \\
    Wan 2.7 & 0.094 & 0.205 & 0.183 & \best{0.214} & \best{0.687} & \best{0.916} & \third{0.307} & \second{0.787} & \third{0.172} & 0.000 & \third{0.312} & \second{0.625} & 0.000 & 0.000 & \second{0.136} & \third{0.410} & \best{0.869} & 0.246 & 0.299 & \third{0.138} \\
    LTX 2.3 & 0.038 & 0.119 & 0.089 & 0.000 & 0.445 & 0.603 & 0.264 & 0.715 & 0.125 & 0.000 & 0.188 & 0.375 & 0.000 & 0.000 & 0.019 & 0.257 & 0.567 & 0.211 & 0.209 & 0.100 \\
    Wan 2.2-LoRA$_{2K}$ & 0.019 & 0.095 & 0.066 & 0.024 & 0.451 & 0.486 & 0.274 & 0.740 & 0.141 & 0.000 & 0.250 & \third{0.500} & 0.000 & 0.000 & 0.019 & 0.252 & 0.488 & 0.219 & 0.196 & 0.113 \\
    Wan 2.2-LoRA$_{7K}$ & 0.038 & 0.148 & 0.101 & 0.024 & 0.487 & 0.535 & 0.276 & 0.734 & 0.141 & 0.000 & 0.250 & \third{0.500} & 0.000 & 0.000 & 0.029 & 0.296 & 0.529 & 0.221 & 0.223 & 0.113 \\
    CosmosPolicy-DefaultCam & \second{0.189} & 0.248 & 0.195 & 0.048 & 0.507 & 0.530 & \second{0.338} & 0.728 & \second{0.203} & 0.000 & \third{0.312} & \second{0.625} & 0.000 & 0.000 & \third{0.117} & 0.359 & 0.545 & \second{0.270} & 0.297 & \second{0.163} \\
    CosmosPolicy-BenchCam & \best{0.264} & \best{0.307} & \third{0.212} & 0.000 & 0.583 & 0.810 & 0.307 & 0.700 & \third{0.172} & 0.000 & \third{0.312} & \second{0.625} & 0.000 & 0.000 & \second{0.136} & \second{0.420} & 0.781 & 0.245 & \third{0.306} & \third{0.138} \\
    \midrule
    \multicolumn{21}{@{}l}{\scriptsize\textit{Videos generated from enhanced instructions}} \\
    Hailuo 2.3 & \third{0.132} & \third{0.281} & \best{0.255} & 0.119 & 0.574 & 0.777 & 0.314 & 0.728 & \third{0.203} & 0.000 & \second{0.406} & \best{0.750} & \second{0.062} & \second{0.062} & 0.117 & 0.410 & 0.773 & \third{0.263} & \best{0.346} & \third{0.175} \\
    Kling 3.0 & 0.113 & \best{0.284} & \second{0.214} & \second{0.214} & 0.631 & 0.586 & \best{0.460} & \third{0.790} & \best{0.344} & 0.000 & 0.250 & 0.375 & \best{0.125} & \best{0.125} & \second{0.146} & \second{0.423} & 0.552 & \best{0.393} & \second{0.325} & \best{0.300} \\
    SeedDance 2.0 & \second{0.151} & \second{0.283} & \third{0.211} & \best{0.262} & \best{0.656} & 0.797 & 0.302 & 0.762 & \second{0.219} & 0.000 & 0.250 & \third{0.500} & 0.000 & 0.000 & \best{0.184} & \best{0.433} & 0.749 & 0.242 & \third{0.317} & \third{0.175} \\
    Veo 3.1 & 0.028 & 0.084 & 0.086 & \third{0.122} & \third{0.635} & \best{0.919} & 0.281 & \second{0.826} & 0.145 & 0.000 & 0.188 & 0.375 & 0.000 & 0.000 & 0.071 & 0.359 & \third{0.830} & 0.223 & 0.250 & 0.115 \\
    Wan 2.2 & 0.038 & 0.154 & 0.094 & 0.095 & 0.545 & 0.633 & 0.304 & 0.738 & 0.172 & 0.000 & 0.188 & 0.375 & 0.000 & 0.000 & 0.058 & 0.316 & 0.592 & 0.243 & 0.218 & 0.138 \\
    Wan 2.7 & 0.094 & 0.225 & 0.154 & \second{0.214} & \second{0.647} & \third{0.852} & \second{0.343} & 0.733 & \third{0.203} & 0.000 & \best{0.438} & \best{0.750} & \best{0.125} & \best{0.125} & \third{0.136} & \third{0.414} & \second{0.836} & \second{0.299} & 0.266 & \second{0.188} \\
    LTX 2.3 & 0.057 & 0.161 & 0.112 & 0.075 & 0.561 & 0.821 & 0.323 & 0.729 & 0.183 & 0.000 & 0.312 & \second{0.625} & 0.000 & 0.000 & 0.059 & 0.331 & 0.789 & 0.255 & 0.230 & 0.145 \\
    Wan 2.2-LoRA$_{2K}$ & 0.057 & 0.149 & 0.091 & 0.119 & 0.550 & 0.604 & \third{0.329} & 0.786 & \third{0.203} & 0.000 & 0.188 & 0.375 & 0.000 & 0.000 & 0.078 & 0.315 & 0.568 & 0.263 & 0.225 & 0.163 \\
    Wan 2.2-LoRA$_{7K}$ & 0.019 & 0.140 & 0.078 & 0.119 & 0.546 & 0.656 & 0.296 & \best{0.865} & 0.172 & 0.000 & 0.188 & 0.375 & 0.000 & 0.000 & 0.059 & 0.311 & 0.611 & 0.237 & 0.229 & 0.138 \\
    CosmosPolicy-DefaultCam & \best{0.170} & 0.233 & 0.178 & 0.000 & 0.561 & 0.663 & 0.276 & 0.769 & 0.141 & 0.000 & 0.188 & 0.375 & 0.000 & 0.000 & 0.087 & 0.363 & 0.617 & 0.221 & 0.292 & 0.113 \\
    CosmosPolicy-BenchCam & \second{0.151} & 0.235 & 0.165 & 0.000 & 0.605 & \second{0.887} & 0.277 & 0.716 & 0.141 & 0.000 & \third{0.375} & \best{0.750} & 0.000 & 0.000 & 0.078 & 0.397 & \best{0.865} & 0.222 & 0.271 & 0.113 \\
    \midrule
    \rowcolor{gray!10}
    \rowcolor{gray!10}
    Rollout Video$^\dagger$ & 0.765 & 0.851 & 0.818 & 0.381 & 0.742 & 0.811 & 0.562 & 0.755 & 0.516 & 0.750 & 0.938 & 1.000 & 0.875 & 0.875 & 0.604 & 0.812 & 0.842 & 0.625 & 0.805 & 0.588 \\
    \rowcolor{gray!18}
    Rollout Video w/ GT Depth$^\ddagger$ & 1.000 & 1.000 & 0.950 & 0.952 & 0.979 & 0.905 & 0.866 & 1.000 & 0.953 & 1.000 & 1.000 & 1.000 & 1.000 & 1.000 & 0.981 & 0.991 & 0.920 & 0.893 & 0.960 & 0.963 \\
    \bottomrule
    \end{tabular}
  }
  \vspace{2pt}
  \begin{minipage}{0.97\textwidth}
    \scriptsize
    \emph{Note.} $^\dagger$ Rollout Video and $^\ddagger$ Rollout Video (w/ GT Depth) serve as oracle/reference bounds. Rankings are computed separately within each instruction setting among generation models only; oracle/reference rows are shaded in gray. Zero-valued entries are not highlighted, even when tied.
  \end{minipage}
  
  \vspace{-1em}
\end{table*}

\section{More Results}
\label{app:more_results}

\begin{figure*}[t]
    \centering
    \includegraphics[width=0.75\textwidth]{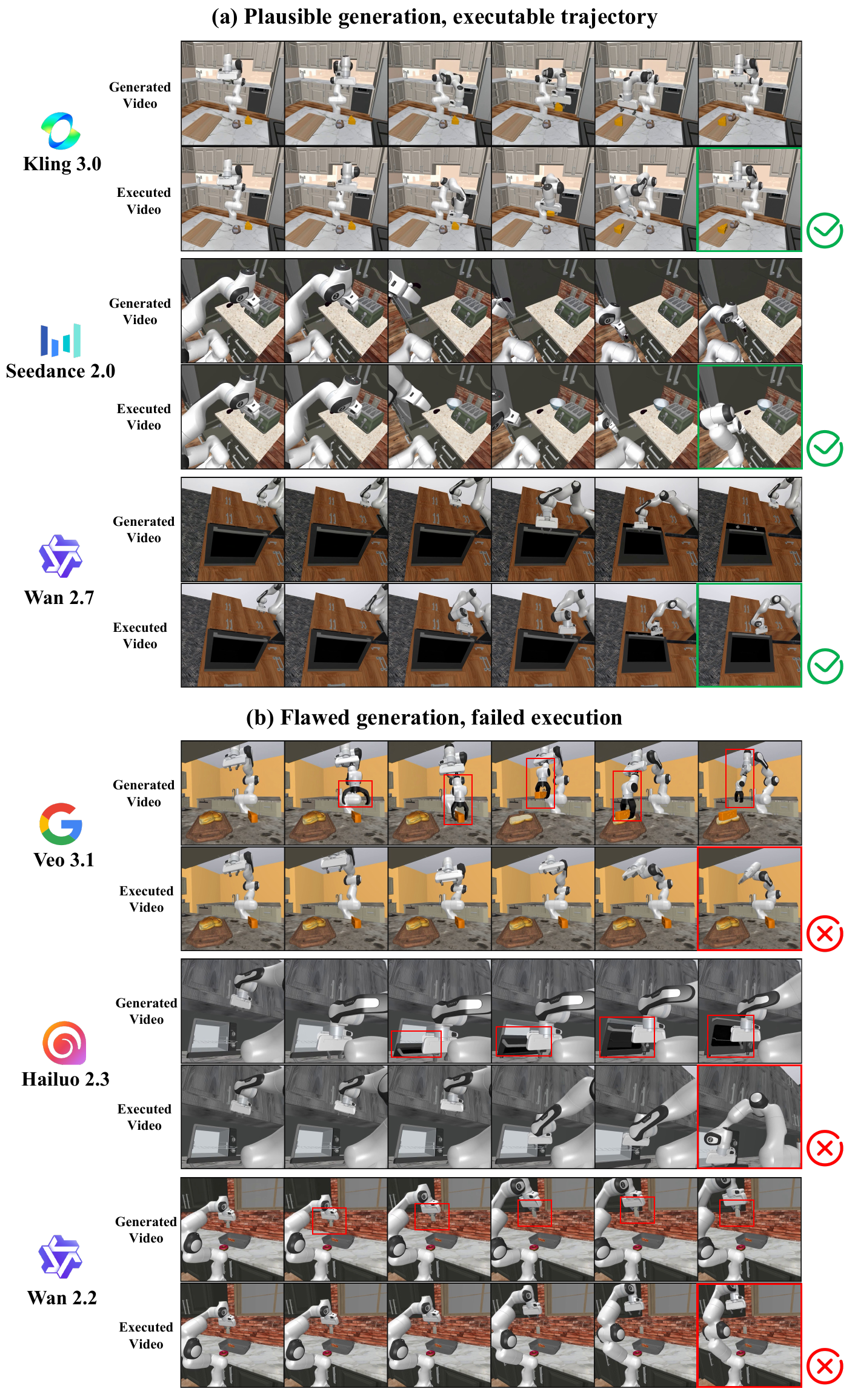}
    \caption{
    \textbf{Qualitative examples of video-to-execution outcomes.}
    Each example shows six temporally aligned frames from the generated video and the recovered execution rollout.
    \textbf{(a)} Successful cases show that visually plausible robot-object motion can be converted into executable trajectories and completed rollouts.
    \textbf{(b)} Failure cases illustrate how generation artifacts, such as inconsistent robot geometry, object-state hallucinations, and unreliable contact cues, propagate through trajectory extraction and lead to failed execution.
    }
    \label{fig:success_failure_cases}
\end{figure*}

Figure~\ref{fig:success_failure_cases} provides additional qualitative examples for our evaluation protocol. For each example, we compare the generated manipulation video with the rollout video obtained after executing the corresponding action trajectory. 

The successful cases show that coherent object motion and stable robot-object interactions can be recovered as executable trajectories. The failure cases illustrate the opposite behavior, where artifacts such as spurious objects or inconsistent contacts introduce visual evidence that cannot be mapped to a valid robot action sequence. 

\end{document}